\documentclass{article}

\usepackage[preprint, nonatbib]{neurips_2021}

\usepackage[utf8]{inputenc} 
\usepackage[T1]{fontenc}    
\usepackage{url}            
\usepackage{booktabs}       
\usepackage{amsfonts}       
\usepackage{nicefrac}       
\usepackage{microtype}      
\usepackage{xcolor}         

\usepackage{amsmath}
\usepackage{amssymb}
\usepackage{float}
\usepackage{multirow}
\usepackage{xcolor}
\usepackage{caption}
\usepackage{subcaption}
\usepackage{program}
\catcode`\|=12\relax
\usepackage{tikz}

\usepackage[pagebackref=true]{hyperref} 
\renewcommand*\backref[1]{\ifx#1\relax \else (Cited on page #1) \fi}

\title{Learning the Travelling Salesperson Problem\\Requires Rethinking Generalization}

\author{%
  Chaitanya K. Joshi$^1$, Quentin Cappart$^2$, Louis-Martin Rousseau$^2$, Thomas Laurent$^3$ \\
  $^1$University of Cambridge, UK \\
  $^2$Ecole Polytechnique de Montréal, Canada \\
  $^3$Loyola Marymount University, USA \\
  \texttt{chaitanya.joshi@cl.cam.ac.uk}
}

\begin{document}

\maketitle

\begin{abstract}

End-to-end training of neural network solvers for graph combinatorial optimization problems such as the Travelling Salesperson Problem (TSP) have seen a surge of interest recently, but remain intractable and inefficient beyond graphs with few hundreds of nodes. 
While state-of-the-art learning-driven approaches for TSP perform closely to classical solvers when trained on trivially small sizes, they are unable to generalize the learnt policy to larger instances at practical scales.
This work presents an end-to-end \textit{neural combinatorial optimization} pipeline that unifies several recent papers in order to identify the inductive biases, model architectures and learning algorithms that promote generalization to instances larger than those seen in training. 
Our controlled experiments provide the first principled investigation into such \textit{zero-shot} generalization, revealing that extrapolating beyond training data requires rethinking the neural combinatorial optimization pipeline, from network layers and learning paradigms to evaluation protocols.
Additionally, we analyze recent advances in deep learning for routing problems through the lens of our pipeline and provide new directions to stimulate future research.\footnote{
Code and datasets: \url{github.com/chaitjo/learning-tsp}}

\end{abstract}


\section{Introduction}
\label{sec:intro}

NP-hard combinatorial optimization problems are the family of integer constrained optimization problems which are intractable to solve optimally at large scales. 
Robust approximation algorithms to popular problems have immense practical applications and are the backbone of modern industries.
Among combinatorial problems, the 2D Euclidean Travelling Salesperson Problem~(TSP) has been the most intensely studied NP-hard graph problem in the Operations Research~(OR) community, with applications in logistics, genetics and scheduling~\cite{lenstra1975some}.
TSP is intractable to solve optimally above thousands of nodes for modern computers~\cite{applegate2006traveling}.
In practice, the Concorde TSP solver~\cite{applegate2006concorde}
uses linear programming with carefully handcrafted heuristics to find solutions up to tens of thousands of nodes, but with prohibitive execution times.\footnote{
The largest TSP solved by Concorde to date has 109,399 nodes with running time of 7.5 months.
}
Besides, the development of problem-specific OR solvers such as Concorde for novel or under-studied problems arising in scientific discovery~\cite{senior2020improved} or computer architecture~\cite{mirhoseini2021graph} requires significant time and specialized knowledge.

An alternate approach by the Machine Learning community is to develop generic learning algorithms which can be trained to solve \textit{any} combinatorial problem directly from problem instances themselves~\cite{vinyals2015pointer, bello2016neural, bengio2018machine}.
Using classical problems such as TSP, Minimum Vertex Cover and Boolean Satisfiability as benchmarks, recent \textit{end-to-end} approaches~\cite{khalil2017learning, selsam2018learning, li2018combinatorial} leverage advances in graph representation learning~\cite{kipf2017semi, gilmer2017neural, velickovic2018graph, battaglia2018relational}
and have shown competitive performance with OR solvers on trivially small problem instances up to few hundreds of nodes.
Once trained, approximate solvers based on Graph Neural Networks (GNNs)
have significantly favorable time complexity than their OR counterparts, making them highly desirable for real-time decision-making problems such as TSP and the associated class of Vehicle Routing Problems (VRPs). 


\begin{figure}[t]
    \centering
    \includegraphics[width=0.95\linewidth]{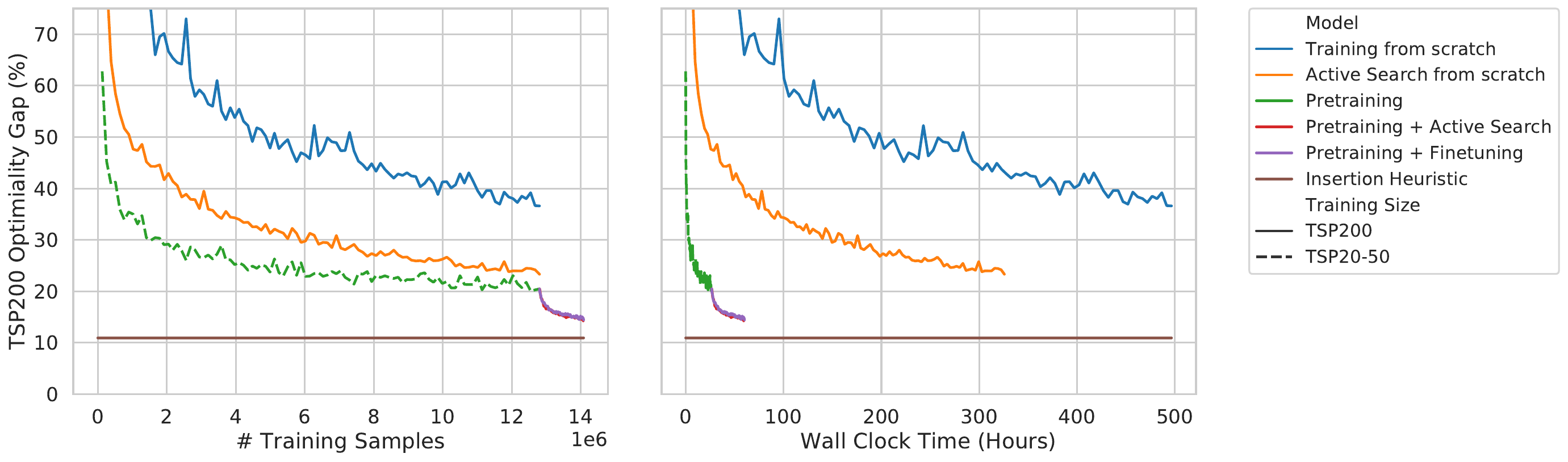}
    \caption{\small
    \textbf{Computational challenges of learning large scale TSP.} We compare three identical autoregressive GNN-based models trained on 12.8~Million TSP instances via reinforcement learning. We plot average optimality gap to the Concorde solver on 1,280 held-out TSP200 instances vs. number of training samples (left) and wall clock time (right) during the learning process.
    Training on large TSP200 from scratch is intractable and sample inefficient.
    Active Search~\cite{bello2016neural}, which learns to directly overfit to the 1,280 held-out samples, further demonstrates the computational challenge of memorizing very few TSP200 instances.
    Comparatively, learning efficiently from trivial TSP20-TSP50 allows models to better generalize to TSP200 in a zero-shot manner, indicating positive knowledge transfer from small to large graphs.
    Performance can further improve via rapid finetuning on 1.28~Million TSP200 instances or by Active Search.
    Within our computational budget, a simple non-learnt \textit{furthest insertion} heuristic still outperforms all models.
    Precise experimental setup is described in \textbf{Appendix~\ref{app:story}}. 
    }
    \label{fig:story}
\end{figure}


\textbf{Motivation\quad}
\label{sec:intro:motivation}
Scaling end-to-end approaches to practical and real-world instances is still an open question~\cite{bengio2018machine} 
as the training phase of state-of-the-art models on large graphs is extremely time-consuming.
For graphs larger than few hundreds of nodes, the gap between GNN-based solvers and simple non-learnt heuristics is especially evident for routing problems like TSP~\cite{kool2018attention, joshi2019efficient}.

As an illustration, Figure~\ref{fig:story} presents the computational challenge of learning TSP on 200-node graphs~(TSP200) in terms of both sample efficiency and wall clock time. 
Surprisingly, it is difficult to outperform a simple insertion heuristic when directly training on 12.8~Million TSP200 samples for 500~hours on university-scale hardware. 

We advocate for an alternative to expensive large-scale training: learning efficiently from trivially small TSP and transferring the learnt policy to larger graphs in a \textit{zero-shot} fashion or via fast finetuning. 
Thus, identifying promising inductive biases, architectures and learning paradigms that enable such zero-shot generalization to large and more complex instances is a key concern for training practical solvers for real-world problems.

\textbf{Contributions\quad}
Towards end-to-end learning of \textit{scale-invariant} TSP solvers, we unify several state-of-the-art architectures and learning paradigms~\cite{nowak2017note,kool2018attention,deudon2018learning,joshi2019efficient} into one experimental pipeline and provide the first principled investigation on zero-shot generalization to large instances.
Our findings suggest that learning scale-invariant TSP solvers requires rethinking the status quo of neural combinatorial optimization to explicitly account for generalization:
\begin{itemize}
    \item The prevalent evaluation paradigm overshadows models' poor generalization capabilities by measuring performance on fixed or trivially small TSP sizes.
    \item Generalization performance of GNN aggregation functions and normalization schemes benefits from explicit redesigns which account for shifting graph distributions, and can be further boosted by enforcing regularities such as constant graph diameters when defining problems using graphs.
    \item Autoregressive decoding enforces a sequential inductive bias which improves generalization over non-autoregressive models, but is costlier in terms of inference time.
    \item Models trained with expert supervision are more amenable to post-hoc search, while reinforcement learning approaches scale better with more computation as they do not rely on labelled data.
\end{itemize}
Our framework and datasets are available online\footnote{\url{https://github.com/chaitjo/learning-tsp}}.
Additionally, we use our pipeline to characterize the recent state-of-the-art in deep learning for routing problems and provide new directions for future research~\cite{chaitanyak2022recentadvancesin}.


\section{Related Work}
\label{sec:related}

\textbf{Neural Combinatorial Optimization\quad}
In a recent survey, Bengio~et~al.~\cite{bengio2018machine} identified three broad approaches to leveraging machine learning for combinatorial optimization problems: 
learning alongside optimization algorithms~\cite{gasse2019exact,cappart2019improving,chalumeau2021seapearl}, 
learning to configure optimization algorithms~\cite{wilder2019melding,ferber2020mipaal},
and end-to-end learning to approximately solve optimization problems, \textit{a.k.a.} neural combinatorial optimization~\cite{vinyals2015pointer,bello2016neural}.

State-of-the-art end-to-end approaches for TSP use Graph Neural Networks (GNNs)~\cite{kipf2017semi,gilmer2017neural,velickovic2018graph, battaglia2018relational} and \textit{sequence-to-sequence} learning~\cite{sutskever2014sequence} to construct approximate solutions directly from problem instances.
Architectures for TSP can be classified as: (1) autoregressive approaches, which build solutions in a step-by-step fashion~\cite{khalil2017learning,deudon2018learning,kool2018attention,ma2019combinatorial,kwon2020pomo,ouyang2021generalization}; and
(2) non-autoregressive models, which produce the solution in one shot~\cite{nowak2017note,nowak2018divide,joshi2019efficient,fu2020generalize,kool2021deep}.
Models can be trained to imitate optimal solvers via supervised learning or by minimizing the length of TSP tours via reinforcement learning \cite{joshi2019learning}. 

Other classical problems tackled by similar architectures include Vehicle Routing~\cite{nazari2018reinforcement,chen2019learning}, Maximum Cut~\cite{khalil2017learning}, Minimum Vertex Cover~\cite{li2018combinatorial}, Boolean Satisfiability~\cite{selsam2018learning,yolcu2019learning}, and Graph Coloring~\cite{huang2019coloring}.
Using TSP as an illustration, we present a unified pipeline for characterizing neural combinatorial optimization architectures in Section~\ref{sec:method}.

Notably, TSP has emerged as a challenging testbed for neural combinatorial optimization.
Whereas generalization to problem instances larger and more complex than those seen in training has at least partially been demonstrated on non-sequential problems such as SAT, MaxCut, and MVC~\cite{li2018combinatorial,selsam2018learning},
the same architectures do not show strong generalization for TSP~\cite{kool2018attention,joshi2019efficient}.

\textbf{Combinatorial Optimization and GNNs\quad}
From the perspective of graph representation learning, algorithmic and combinatorial problems have recently been used to characterize the expressive power of GNNs~\cite{sato2019approximation, cappart2021combinatorial}.
An emerging line of work on learning to execute graph algorithms~\cite{velickovic2019neural, velivckovic2021neural} has lead to the development of provably more expressive GNNs~\cite{corso2020principal} and improved understanding of their generalization capability~\cite{xu2019can,xu2020neural}.
Towards tackling realistic and large-scale combinatorial problems, this paper aims to quantify the limitations of prevalent GNN architectures and learning paradigms via zero-shot generalization to problems larger than those seen during training.

\textbf{Novel Applications\quad}
Advances on classical combinatorial problems have shown promising results in downstream applications to novel or under-studied optimization problems in the physical sciences~\cite{gomez2018automatic,senior2020improved} and computer architecture~\cite{mao2019learning,paliwal2019regal,mirhoseini2021graph}, where the development of exact solvers is expensive and intractable.
For example, autoregressive architectures provide a strong inductive bias for device placement optimization problems~\cite{mirhoseini2017device,zhou2019gdp}, while non-autoregressive models~\cite{bresson2019two} are competitive with autoregressive approaches~\cite{jin2018junction,you2018graph} for molecule generation tasks.


\section{Neural Combinatorial Optimization Pipeline}
\label{sec:method}

\begin{figure}
     \centering
     \begin{subfigure}[b]{\textwidth}
        \begin{tikzpicture}
    \definecolor{c1}{RGB}{0,63,92}
    \definecolor{c2}{RGB}{88,80,141}
    \definecolor{c3}{RGB}{188,80,144}
    \definecolor{c4}{RGB}{255,99,97}
    \definecolor{c5}{RGB}{255,166,0}
    
    \tikzset{line/.style={draw,semithick}}
    \tikzset{arrow/.style={line,->,>=stealth}}
    \tikzset{box/.style={line,align=center,text width=2cm,inner sep=1pt,minimum height=1.0cm,rounded corners=8pt}}
    \tikzset{bg1/.style={shading=axis,left color=c1!30!white,right color=c1!30!white,shading angle=45}}
    \tikzset{bg2/.style={shading=axis,left color=c2!45!white,right color=c2!45!white,shading angle=45}}
    \tikzset{bg3/.style={shading=axis,left color=c3!60!white,right color=c3!60!white,shading angle=45}}
    \tikzset{bg4/.style={shading=axis,left color=c4!75!white,right color=c4!75!white,shading angle=45}}
    \tikzset{bg5/.style={shading=axis,left color=c5!90!white,right color=c5!90!white,shading angle=45}}
    
    \node[box,bg1] (1) at (-6,0) {Problem Definition};
    \node[box,bg2] (2) at (-3.5,0) {Graph Embedding};
    \node[box,bg3] (3) at (-1,0) {Solution Decoding};
    \node[box,bg4] (4) at (1.5,0) {Solution Search};
    \node[box,bg5] (5) at (4,0) {Policy Learning};
    
    \path[arrow] (1) to (2);
    \path[arrow] (2) to (3);
    \path[arrow] (3) to (4);
    \path[arrow] (4) to (5);
    
\end{tikzpicture}
        \caption{\small Neural combinatorial optimization pipeline in stages.}
     \end{subfigure}
     \begin{subfigure}[b]{\textwidth}
         \centering
         \includegraphics[width=0.6\textwidth]{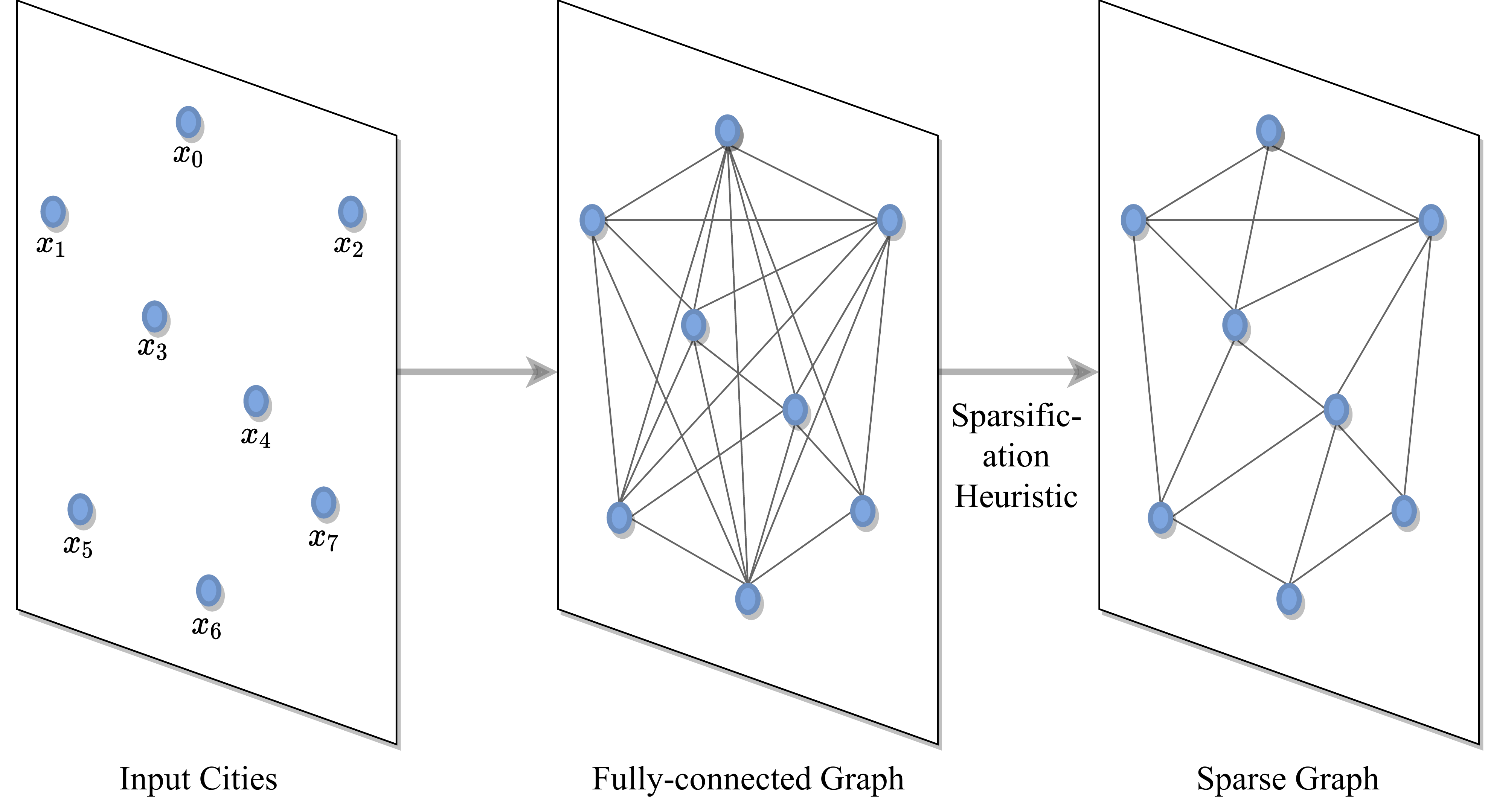}
         \caption{\small \textbf{Problem Definition:} TSP is formulated via a fully-connected graph of cities/nodes. The graph can be sparsified via heuristics such as $k$-nearest neighbors.}
     \end{subfigure}
     \newline
     \begin{subfigure}[b]{\textwidth}
         \centering
         \includegraphics[width=0.6\textwidth]{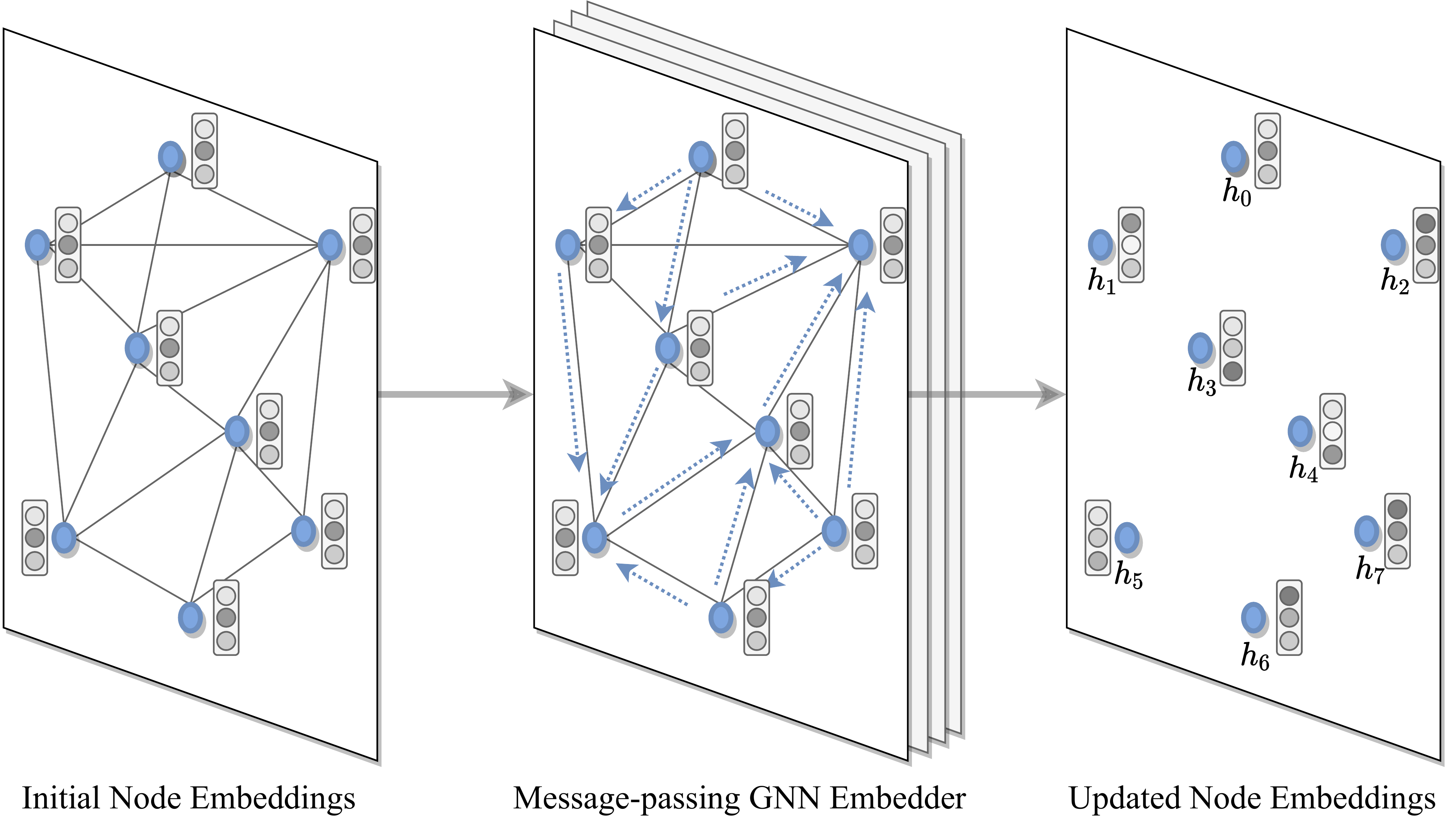}
         \caption{\small \textbf{Graph Embedding:} Embeddings for each graph node are obtained using a Graph Neural Network encoder, which builds local structural features. 
        }
     \end{subfigure}
     \newline
     \begin{subfigure}[b]{\textwidth}
         \centering
         \includegraphics[width=0.65\textwidth]{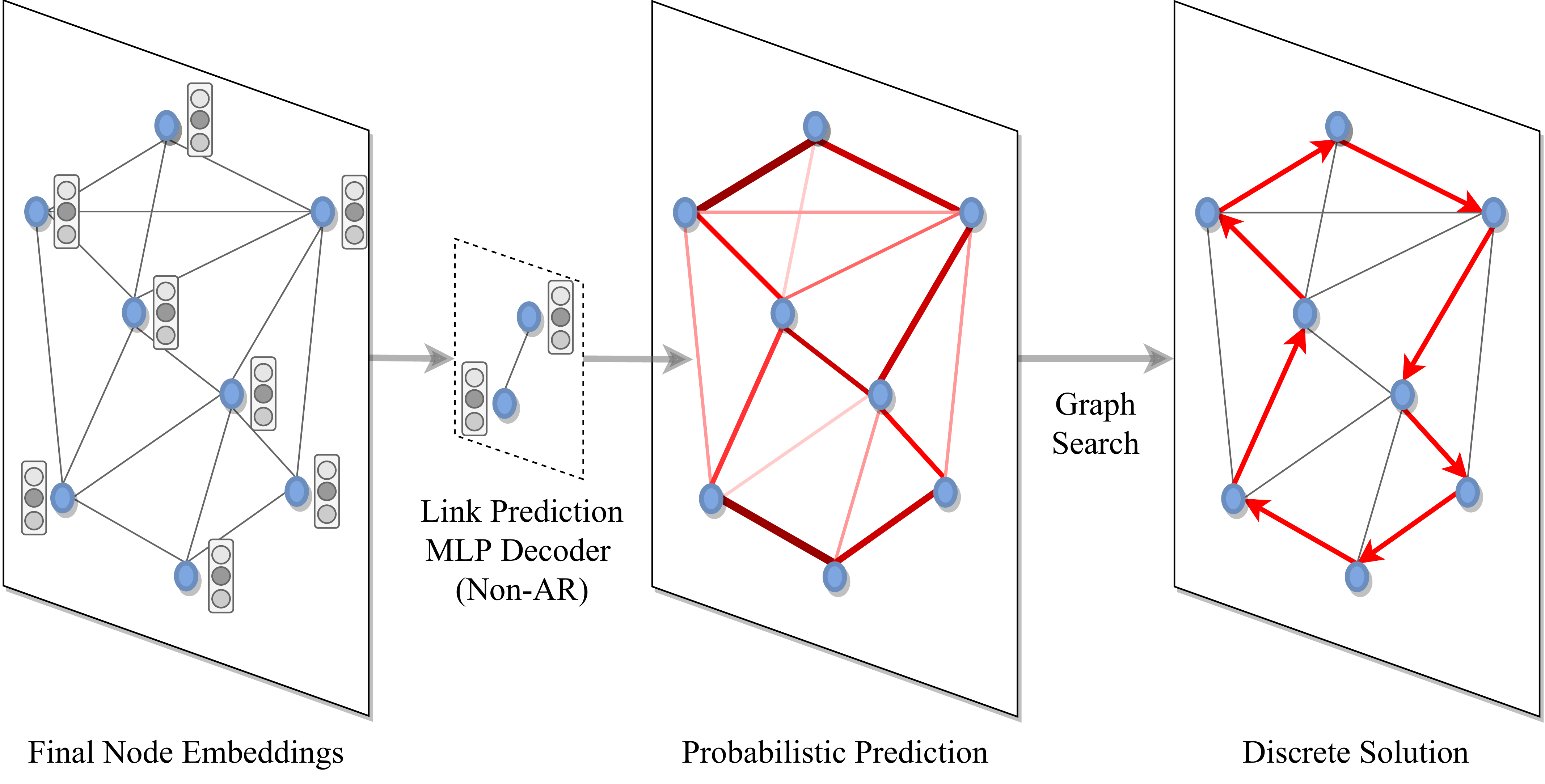}
         \caption{\small \textbf{Solution Decoding \& Search:} Probabilities are assigned to each node for belonging to the solution set, either independent of one-another (\textit{i.e.} Non-autoregressive decoding) or conditionally through graph traversal (\textit{i.e.} Autoregressive decoding).
         The predicted probabilities are converted into discrete decisions through classical graph search techniques such as greedy search or beam search.}
     \end{subfigure}
     \caption{\small {\bf End-to-end neural combinatorial optimization pipeline:} The entire model in trained end-to-end via imitating an optimal solver (\textit{i.e.} supervised learning) or through minimizing a cost function (\textit{i.e.} reinforcement learning).
     }
     \label{fig:pipeline}
\end{figure}
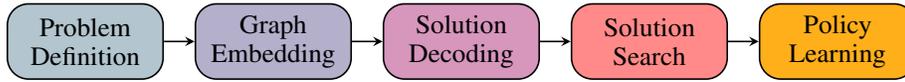
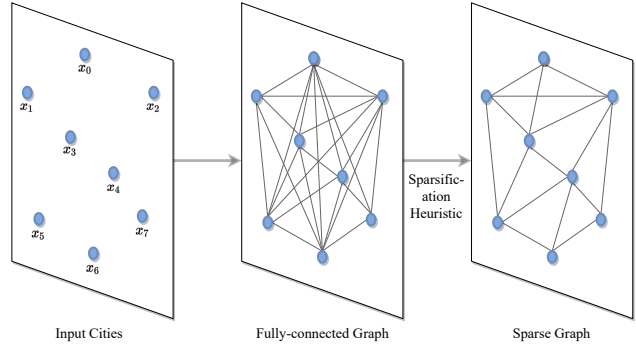
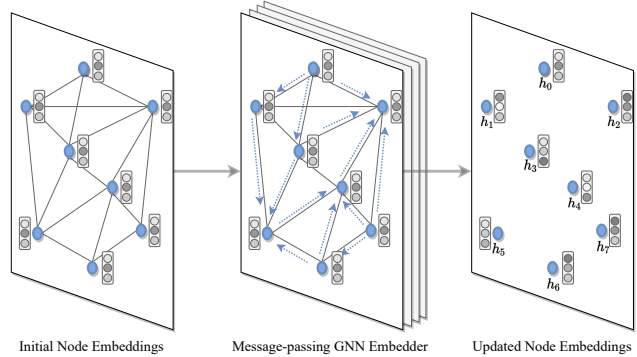
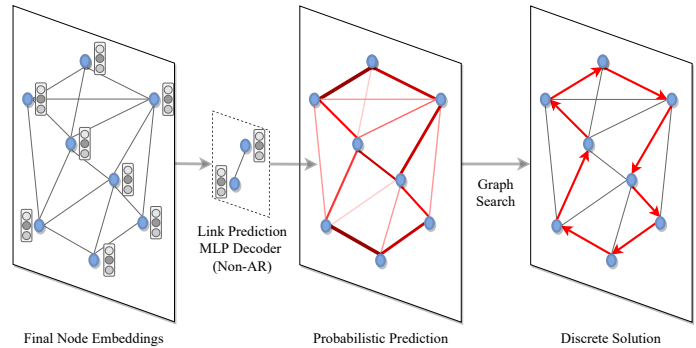

NP-hard problems can be formulated as sequential decision making tasks on graphs due to their highly structured nature. 
Towards a controlled study of neural combinatorial optimization for TSP, we unify recent ideas~\cite{nowak2017note,kool2018attention,deudon2018learning,joshi2019efficient} via a five stage end-to-end pipeline illustrated in Figure~\ref{fig:pipeline}.
Our discussion focuses on TSP, but the pipeline presented is generic and can be extended to characterize modern architectures for several NP-hard graph problems.


\subsection{Problem Definition}

The 2D Euclidean TSP is defined as follows: \textit{``Given a set of cities and the distances between each pair of cities, what is the shortest possible route that visits each city and returns to the origin city?"}
Formally, given a fully-connected input graph of $n$ cities (nodes) in the two dimensional unit square $S = \{x_i\}_{i=1}^n$ where each $x_i \in {[0,1]}^2$, we aim to find a permutation of the nodes $\pi$, termed a tour, that visits each node once and has the minimum total length, defined as:
\begin{equation}
L(\pi | s) = \|x_{\pi_n} - x_{\pi_1}\|_2 + \sum_{i=1}^{n-1} \|x_{\pi_i} - x_{\pi_{i+1}}\|_2,
\end{equation}
where $\|\cdot\|_2$ denotes the $\ell_2$ norm. 

\textbf{Graph Sparsification\quad}
Classically, TSP is defined on fully-connected graphs, see Figure~\ref{fig:pipeline}(b). 
Graph sparsification heuristics based on $k$-nearest neighbors aim to reduce TSP graphs, enabling models to scale up to large instances where pairwise computation for all nodes is intractable~\cite{khalil2017learning} or learn faster by reducing the search space~\cite{joshi2019efficient}.
Notably, problem-specific graph reduction techniques have proven effective for  out-of-distribution generalization to larger graphs for other NP-hard problems such as MVC and SAT~\cite{li2018combinatorial}.

\textbf{Fixed size vs. variable size graphs\quad}
Most work on learning for TSP has focused on training with a fixed graph size~\cite{kool2018attention, joshi2019efficient}, likely due to ease of implementation.
Learning from multiple graph sizes naturally enables better generalization within training size ranges, but its impact on generalization to larger TSP instances remains to be analyzed.


\subsection{Graph Embedding}
\label{sec:method:encoder}

A Graph Neural Network (GNN) encoder computes $d$-dimensional representations for each node in the input TSP graph, see Figure~\ref{fig:pipeline}(c).
At each layer, nodes gather features from their neighbors to represent local graph structure via recursive message passing~\cite{gilmer2017neural}.
Stacking $L$ layers allows the network to build representations from the $L$-hop neighborhood of each node.
Let $h_i^{\ell}$ and $e_{ij}^{\ell}$ denote respectively the node and edge feature at layer $\ell$ associated with node $i$ and edge $ij$. 
We define the feature at the next layer via an \textit{anisotropic} message passing scheme using an edge gating mechanism~\cite{bresson2018experimental}:
\begin{eqnarray}
h_i^{\ell+1} &=& h_i^{\ell} + \text{ReLU} \Big( \textsc{Norm} \Big( U^\ell h_i^{\ell} + \textsc{Aggr}_{j \in \mathcal{N}_i} \Big( \sigma(e_{ij}^{\ell}) \odot V^\ell h_j^{\ell} \Big) \Big)\Big) \ , \label{eqn:gnn-node} \\
e_{ij}^{\ell+1} &=& e^\ell_{ij} + \text{ReLU}\Big( \textsc{Norm} \Big( A^\ell e_{ij}^{\ell} + B^\ell h^{\ell}_i + C^\ell h^{\ell}_j  \Big)\Big) , \label{eqn:gnn-edge}
\end{eqnarray}
where $U^{\ell}, V^{\ell}, A^{\ell}, B^{\ell}, C^{\ell} \in \mathbb{R}^{d \times d}$ are learnable parameters, \textsc{Norm} denotes the normalization layer (BatchNorm~\cite{ioffe2015batch}, LayerNorm~\cite{ba2016layer}), \textsc{Aggr} represents the neighborhood aggregation function (\textsc{Sum}, \textsc{Mean} or \textsc{Max}), $\sigma$ is the sigmoid function, and $\odot$ is the Hadamard product.
As inputs $h_i^{\ell=0}$ and $e_{ij}^{\ell=0}$, we use $d$-dimensional linear projections of the node coordinate $x_i$ and the euclidean distance $\|x_i - x_j\|_2$, respectively.

\textbf{Anisotropic Aggregation\quad}
We make the aggregation function anisotropic or directional via a dense attention mechanism which scales the neighborhood features $h_j, \forall j \in \mathcal{N}_i,$ using edge gates $\sigma(e_{ij})$.
Anisotropic and attention-based GNNs such as Graph Attention Networks~\cite{velickovic2018graph}, Transformers~\cite{vaswani2017attention, joshi2020transformers}, and Gated Graph ConvNets~\cite{bresson2018experimental} have been shown to outperform isotropic Graph ConvNets~\cite{kipf2017semi} across several challenging domains~\cite{dwivedi2020benchmarking}, including TSP~\cite{kool2018attention, joshi2019efficient}.


\subsection{Solution Decoding}
\label{sec:method:decoder}

\textbf{Non-autoregressive Decoding (NAR)\quad}
Consider TSP as a link prediction task: each edge may belong/not belong to the optimal TSP solution independent of one another~\cite{nowak2017note}.
We define the edge predictor as a two layer MLP on the node embeddings produced by the final GNN encoder layer $L$, following Joshi~et~al.~\cite{joshi2019efficient}, see Figure~\ref{fig:pipeline}(d).
For adjacent nodes $i$ and $j$, we compute the unnormalized edge logits:
\begin{equation}
\hat p_{ij} = W_2 \Big( \text{ReLU} \; \big( W_1 \big( \left[ \; h_{G}, \; h_{i}^{L}, \; h_{j}^{L} \; \right] \big) \big) \Big), \;\ \text{where} \;\ h_{G} = \frac{1}{n} \sum_{i=0}^{n} h_i^{L},
\label{eqn:nar-dec}
 \end{equation}
$W_1 \in \mathbb{R}^{3d \times d}, W_2 \in \mathbb{R}^{d \times 2}$, and $[\cdot,\cdot,\cdot]$ is the concatenation operator.
The logits $\hat p_{ij}$ are converted to probabilities over each edge $p_{ij}$ via a softmax.

\textbf{Autoregressive Decoding (AR)\quad}
Although NAR decoders are fast as they produce predictions in one shot, they ignore the sequential ordering of TSP tours.
Autoregressive decoders, based on attention~\cite{deudon2018learning,kool2018attention} or recurrent neural networks~\cite{vinyals2015pointer,ma2019combinatorial}, explicitly model this sequential inductive bias through step-by-step graph traversal. 
We follow the attention decoder from Kool~et~al.~\cite{kool2018attention}, which starts from a random node and outputs a probability distribution over its neighbors at each step.
Greedy search is used to perform the traversal over $n$ time steps and masking enforces constraints such as not visiting previously visited nodes.

At time step $t$ at node $i$, the decoder builds a context $\hat h_{i}^{C}$ for the partial tour $\pi'_{t'}$, generated at time $t' < t$, by packing together the graph embedding $h_{G}$ and the embeddings of the first and last node in the partial tour: $\hat h_{i}^{C} = W_{C} \left[ \; h_{G}, \; h^{L}_{\pi'_{t-1}}, \; h^{L}_{\pi'_1} \; \right],$
where $W_{C} \in \mathbb{R}^{3d \times d}$ and learned placeholders are used for $h^{L}_{\pi'_{t-1}}$ and $h^{L}_{\pi'_1}$ at $t = 1$.
The context $\hat h_{i}^{C}$ is then refined via a standard Multi-Head Attention~(MHA) operation~\cite{vaswani2017attention} over the node embeddings:
\begin{equation}
h_{i}^{C} = \text{MHA} \Big( Q=\hat h_{i}^{C}, K=\{h^{L}_1,\dots,h^{L}_n\}, V=\{h^{L}_1,\dots,h^{L}_n\} \Big),
\label{eqn:ar-mha}
\end{equation}
where $Q, K, V$ are inputs to the $M$-headed MHA ($M=8$). 
The unnormalized logits for each edge $e_{ij}$ are computed via a final attention mechanism  between the context $h_{i}^{C}$ and the embedding $h_{j}$: 
\begin{equation}
    \hat p_{ij} = \begin{cases}
        C \cdot \tanh \left( \frac{ \left( W_{Q} h_{i}^{C} \right)^{T} \cdot \left( W_{K} h^{L}_{j} \right) }{\sqrt{d}} \right) & \text{if } j \neq \pi_{t'} \quad \forall t' < t \\
        -\infty & \text{otherwise}.
    \end{cases}
    \label{eqn:ar-logits}
\end{equation}
The $\tanh$ is used to maintain the value of the logits within $[-C, C]$ ($C=10$)~\cite{bello2016neural}.
The logits $\hat p_{ij}$ at the current node $i$ are converted to probabilities $p_{ij}$ via a softmax over all edges.

\textbf{Inductive Biases\quad}
NAR approaches, which make predictions over edges independently of one-another, have shown strong out-of-distribution generalization for non-sequential problems such as SAT and MVC~\cite{li2018combinatorial}.
On the other hand, AR decoders come with the sequential/tour constraint built-in and are the default choice for routing problems~\cite{kool2018attention}.
Although both approaches have shown close to optimal performance on fixed and small TSP sizes under different experimental settings, it is important to fairly compare which inductive biases are most useful for generalization.


\subsection{Solution Search}

\textbf{Greedy Search\quad}
For AR decoding, the predicted probabilities at node $i$ are used to select the edge to travel along at the current step via sampling from the probability distribution $p_i$ or greedily selecting the most probable edge $p_{ij}$, \textit{i.e.} greedy search.
Since NAR decoders directly output probabilities over all edges independent of one-another, we can obtain valid TSP tours using greedy search to traverse the graph starting from a random node and masking previously visited nodes.
Thus, the probability of a partial tour $\pi'$ can be formulated as $p(\pi') = \prod_{j' \sim i' \in \pi'} p_{i'j'}$, where each node $j'$ follows node $i'$.

\textbf{Beam Search and Sampling\quad}
During inference, we can increase the capacity of greedy search via limited width breadth-first beam search, which maintains the $b$ most probable tours during decoding. 
Similarly, we can sample $b$ solutions from the learnt policy and select the shortest tour among them.
Naturally, searching longer, with more sophisticated techniques, or sampling more solutions allows trading off run time for solution quality.
However, it has been noted that using large $b$ for search/sampling or local search during inference may overshadow an architecture's inability to generalize~\cite{francois2019how}.
To better understand generalization, we focus on using greedy search and beam search/sampling with small $b = 128$. 


\subsection{Policy Learning}

\textbf{Supervised Learning\quad}
Models can be trained end-to-end via imitating an optimal solver at each step (\textit{i.e.} supervised learning).
For models with NAR decoders, the edge predictions are linked to the ground-truth TSP tour by minimizing the binary cross-entropy loss for each edge~\cite{joshi2019efficient}.
For AR architectures, at each step, we minimize the cross-entropy loss between the predicted probability distribution over all edges leaving the current node and the next node from the groundtruth tour, following Vinyals~et~al.~\cite{vinyals2015pointer}.
We use teacher-forcing to stabilize training~\cite{williams1989learning}.

\textbf{Reinforcement Learning\quad}
Reinforcement learning is a elegant alternative in the absence of groundtruth solutions, as is often the case for understudied combinatorial problems.
Models can be trained by minimizing problem-specific cost functions (the tour length in the case of TSP) via policy gradient algorithms~\cite{bello2016neural,kool2018attention} or Q-Learning~\cite{khalil2017learning}.
We focus on policy gradient methods due to their simplicity, and define the loss for an instance $s$ parameterized by the model $\theta$ as $\mathcal{L}(\theta | s) = \mathbb{E}_{p_{\theta}(\pi | s)}\left[L(\pi)\right]$, the expectation of the tour length $L(\pi)$, where $p_{\theta}(\pi | s)$ is the probability distribution from which we sample to obtain the tour $\pi|s$.
We use the REINFORCE gradient estimator~\cite{williams1992simple} to minimize $\mathcal{L}$:
\begin{equation}
\label{eq:reinforce_baseline}
	\nabla \mathcal{L}(\theta | s) = \mathbb{E}_{p_{\theta}(\pi | s)}\left[\left(L(\pi) - b(s)\right) \nabla \log p_{\theta}(\pi | s)\right],
\end{equation}
where the baseline $b(s)$ reduces gradient variance.
Our experiments compare standard critic network baselines~\cite{bello2016neural,deudon2018learning} and the greedy rollout baseline proposed by Kool~et~al.~\cite{kool2018attention}.



\section{Experimental Setup}
\label{sec:setup}

We design controlled experiments to probe the unified pipeline described in Section~\ref{sec:method} in order to identify inductive biases, architectures and learning paradigms that promote zero-shot generalization.
We focus on learning efficiently from small problem instances (TSP20-50) and measure generalization to a wider range of sizes, including large instances which are intractable to learn from (\textit{e.g.} TSP200).
Each experiment starts with a `base' model configuration and ablates the impact of a specific component of the five-stage pipeline.
We aim to fairly compare state-of-the-art ideas in terms of model capacity and training data, and expect models with good inductive biases for TSP to: 
(1)~learn trivially small TSPs without hundreds of millions of training samples and model parameters; and
(2)~generalize reasonably well across smaller and larger instances than those seen in training.

To quantify `good' generalization, we additionally evaluate our models against a simple, non-learnt \textit{furthest insertion} heuristic baseline, 
which constructively builds a partial tour $\pi'$ by inserting node $i$ between tour nodes $j_1, j_2 \in \pi'$ such that the distance from node $i$ to its nearest node $j_1$ is maximized. 
Kool~et~al.~\cite{kool2018attention} provide a detailed description of insertion heuristic baselines.

\textbf{Training Datasets\quad}
We perform ablation studies of each component of the pipeline by training on variable TSP20-50 graphs for rapid experimentation.
We also compare to learning from fixed graph sizes up to TSP100. 
Each TSP instance consist of $n$ nodes sampled uniformly in the unit square $S = \{x_i\}_{i=1}^n$ and $x_i \in {[0,1]}^2$.
In the supervised learning paradigm, we generate a training set of 1,280,000 TSP samples and groundtruth tours using the optimal Concorde solver as an oracle.
Models are trained using the Adam optimizer for 10 epochs with a batch size of 128 and a fixed learning rate $1e-4$.
For reinforcement learning, models are trained for 100 epochs on 128,000 TSP samples which are randomly generated for each epoch (without optimal solutions) with the same batch size and learning rate.
Thus, both learning paradigms see 12,800,000 TSP samples in total.
Considering that TSP20-50 are trivial in terms of complexity as they can be solved by simpler non-learnt heuristics, training good solvers at this scale should ideally not require billions of instances.

\textbf{Model Hyperparameters\quad}
For models with AR decoders, we use 3 GNN encoder layers followed by the attention decoder head, setting hidden dimension $d=128$.
For NAR models, we use the same hidden dimension and opt for 4 GNN encoder layers followed by the edge predictor. 
This results in approximately 350,000 trainable parameters for each model, irrespective of decoder type.
Unless specified, most experiments use our best model configuration: AR decoding scheme and Graph ConvNet encoder with \textsc{Max} aggregation and BatchNorm (with batch statistics).
All models are trained via supervised learning except when comparing learning paradigms.

\textbf{Evaluation\quad}
We compare models on a held-out test set of 25,600 TSPs, consisting of 1,280 samples each of TSP10, TSP20, $\dots$, TSP200.
Our evaluation metric is the optimality gap \textit{w.r.t.} the Concorde solver, \textit{i.e.} the average percentage ratio of predicted tour lengths relative to optimal tour lengths.
To compare design choices among identical models, we plot line graphs of the optimality gap as TSP size increases (along with a 99\%-ile confidence interval) using beam search with a width of 128. 
Compared to previous work which evaluated on fixed problem sizes, our protocol identifies not only those models that perform well on training sizes, but also those that generalize better than non-learnt heuristics for large instances which are intractable to train on.


\begin{figure}[t!]
\centering
\begin{minipage}{.46\textwidth}
\centering
    \includegraphics[width=0.98\linewidth]{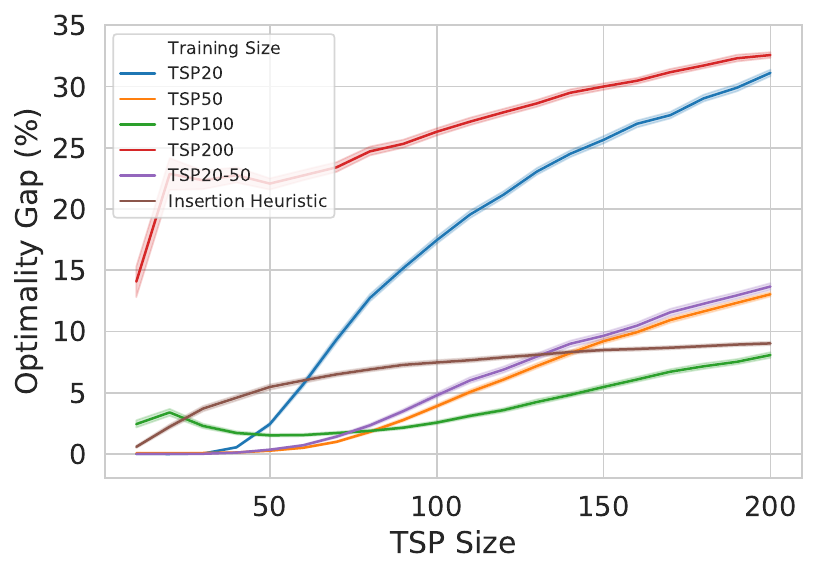}
    \caption{\small\textbf{Learning from various TSP sizes.} The prevalent protocol of evaluation on training sizes overshadows brittle out-of-distribution performance to larger and smaller graphs.}
    \label{fig:fixed_vs_var}
\end{minipage}
\quad
\begin{minipage}{.48\textwidth}
\centering
    \includegraphics[width=0.98\linewidth]{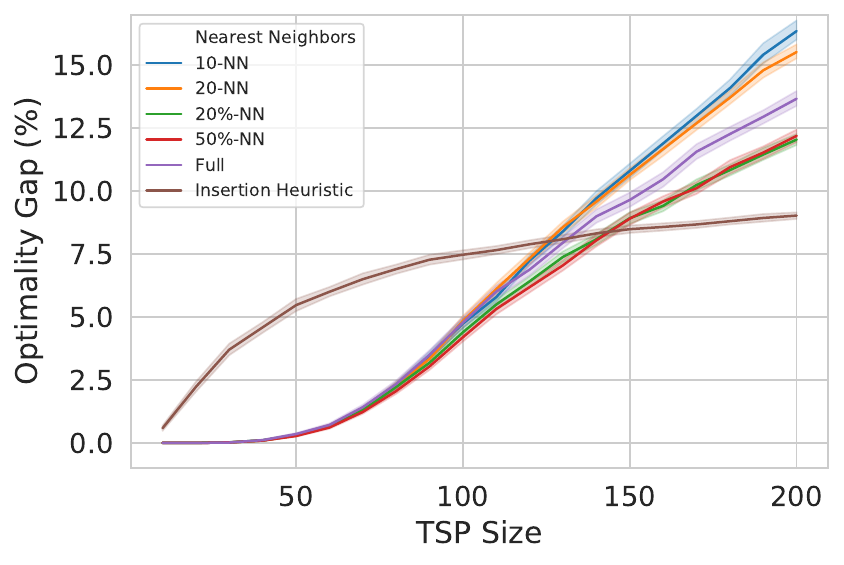}
    \caption{\small\textbf{Impact of graph sparsification.} Maintaining a constant graph diameter across TSP sizes leads to better generalization on larger problems than using full graphs.}
    \label{fig:full_vs_knn}
\end{minipage}
\end{figure}


\section{Results}
\label{sec:experiments}

\subsection{Does learning from variable sizes help generalization?}

We train five identical models on fully connected graphs of instances from TSP20, TSP50, TSP100, TSP200 and variable TSP20-50. 
The line plots of optimality gap across TSP sizes in Figure~\ref{fig:fixed_vs_var} indicates that learning from variable TSP sizes helps models retain performance across the range of graph sizes seen during training (TSP20-50).
Variable graph training compared to training solely on the maximum sized instances (TSP50) leads to marginal gains on small instances but, somewhat counter-intuitively, does not enable better generalization to larger problems.
Learning from small TSP20 is unable to generalize to large sizes while TSP100 models generalize poorly to trivially easy sizes, suggesting that the prevalent protocol of evaluation on training sizes~\cite{kool2018attention,joshi2019efficient} overshadows brittle out-of-distribution performance.

Training on TSP200 graphs is intractable within our computational budget, see Figure~\ref{fig:story}.
TSP100 is the only model which generalizes better to large TSP200 than the non-learnt baseline.
However, training on TSP100 can also be prohibitively expensive: one epoch takes approximately 8 hours (TSP100) vs. 2 hours (TSP20-50) (details in Appendix~\ref{app:hardware}).
For rapid experimentation, we train efficiently on variable TSP20-50 for the rest of our study.


\subsection{What is the best graph sparsification heuristic?}

Figure~\ref{fig:full_vs_knn} compares full graph training to the following heuristics:
(1) \textbf{Fixed node degree} across graph sizes, via connecting each node in TSP$n$ to its $k$-nearest neighbors, enabling GNN encoder layers to specialize to constant degree $k$; and
(2) \textbf{Fixed graph diameter} across graph sizes, via connecting each node in TSP$n$ to its $n\times k\%$-nearest neighbors, ensuring that the same number of message passing steps are required to diffuse information across both small and large graphs.

Although both sparsification techniques lead to faster convergence on training instance sizes (not shown), we find that only approach~(2) leads to better generalization on larger problems than using full graphs.
Consequently, all further experiments use approach~(2) to operate on sparse $20\%$-nearest neighbors graphs.
Our results also suggest that developing more principled problem definition and graph reduction techniques beyond simple $k$-nearest neighbors for augmenting learning-based approaches may be a promising direction.


\subsection{What is the relationship between GNN aggregation functions and normalization layers?}
\label{sec:experiments:aggr-norm}

\begin{figure}[t!]
\centering
\begin{minipage}{.48\textwidth}
\centering
    \includegraphics[width=0.98\linewidth]{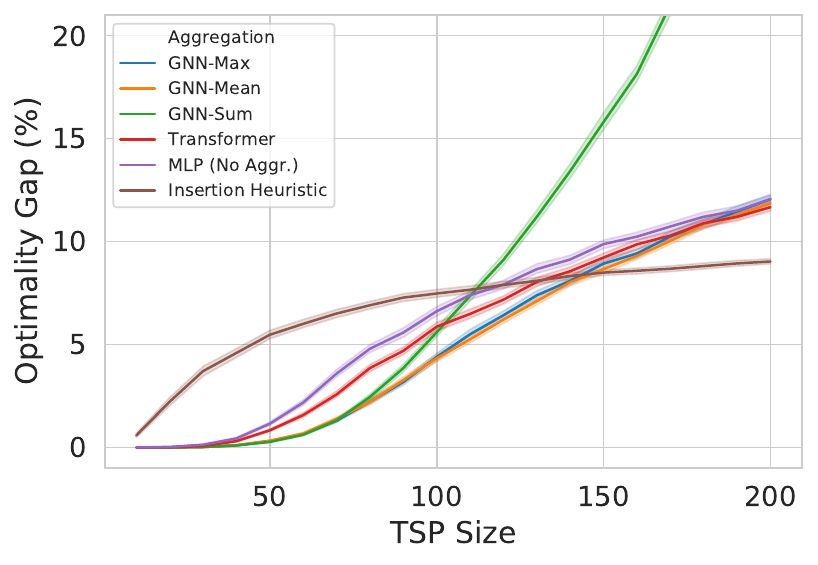}
    \caption{\small\textbf{Impact of GNN aggregation functions.} For larger graphs, aggregators that are agnostic to node degree (\textsc{Mean}, \textsc{Max}) are able to outperform theoretically more expressive aggregators.}
    \label{fig:gnn_aggregation}
\end{minipage}
\quad
\begin{minipage}{.48\textwidth}
\centering
    \includegraphics[width=0.98\linewidth]{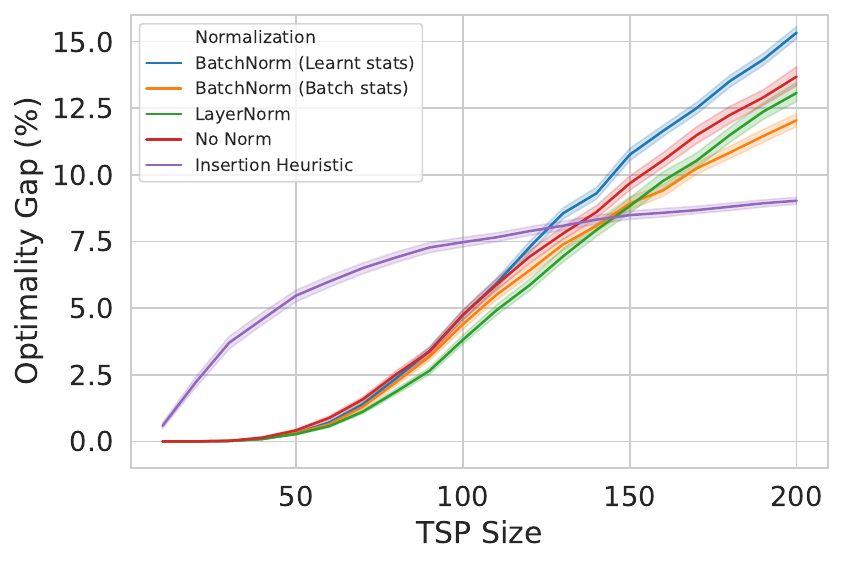}
    \caption{\small\textbf{Impact of normalization schemes.} Modifying BatchNorm to account for changing graph statistics across graph sizes leads to better generalization.}
    \label{fig:gnn_norm}
\end{minipage}
\end{figure}

In Figure~\ref{fig:gnn_aggregation}, we compare identical models with anisotropic \textsc{Sum}, \textsc{Mean} and \textsc{Max} aggregation functions.
As baselines, we consider the Transformer encoder on full graphs~\cite{deudon2018learning,kool2018attention} as well as a structure-agnostic MLP on each node, which can be instantiated by not using any aggregation function in Eq.\eqref{eqn:gnn-node}, \textit{i.e.} $h_i^{\ell+1} = h_i^{\ell} + \text{ReLU} \left( \textsc{Norm} \left( U^\ell h_i^{\ell} \right) \right)$.

We find that the choice of GNN aggregation function does not have an impact when evaluating models within the training size range TSP20-50.
As we tackle larger graphs, GNNs with aggregation functions that are agnostic to node degree (\textsc{Mean} and \textsc{Max}) are able to outperform Transformers and MLPs.
Importantly, the theoretically more expressive \textsc{Sum} aggregator~\cite{xu2018powerful} generalizes worse than structure-agnostic MLPs, as it cannot handle the distribution shift in node degree and neighborhood statistics across graph sizes, leading to unstable or exploding node embeddings~\cite{velickovic2019neural}.
We use the \textsc{Max} aggregator in further experiments, as it generalizes well for both AR and NAR decoders.


We also experiment with the following normalization schemes:
(1) standard BatchNorm which learns mean and variance from training data, as well as (2) BatchNorm with batch statistics; and (3) LayerNorm, which normalizes at the embedding dimension instead of across the batch.
Figure~\ref{fig:gnn_norm} indicates that BatchNorm with batch statistics and LayerNorm are able to better account for changing statistics across different graph sizes.
Standard BatchNorm generalizes worse than not doing any normalization, thus our other experiments use BatchNorm with batch statistics.

We further dissect the relationship between graph representations and normalization in Appendix~\ref{app:graph-and-norm}, confirming that poor performance on large graphs can be explained by unstable representations due to the choice of aggregation and normalization schemes.
Using \textsc{Max} aggregators and BatchNorm with batch statistics are temporary hacks to overcome the failure of the current architectural components.
Overall, our results suggest that inference beyond training sizes will require the development of expressive GNN mechanisms that are able to leverage global graph topology~\cite{garg2020generalization} while being invariant to distribution shifts in terms of node degree and other graph statistics~\cite{levie2019transferability}.



\begin{figure}[t!]
\centering
\begin{minipage}{.48\textwidth}
\centering
    \includegraphics[width=0.98\linewidth]{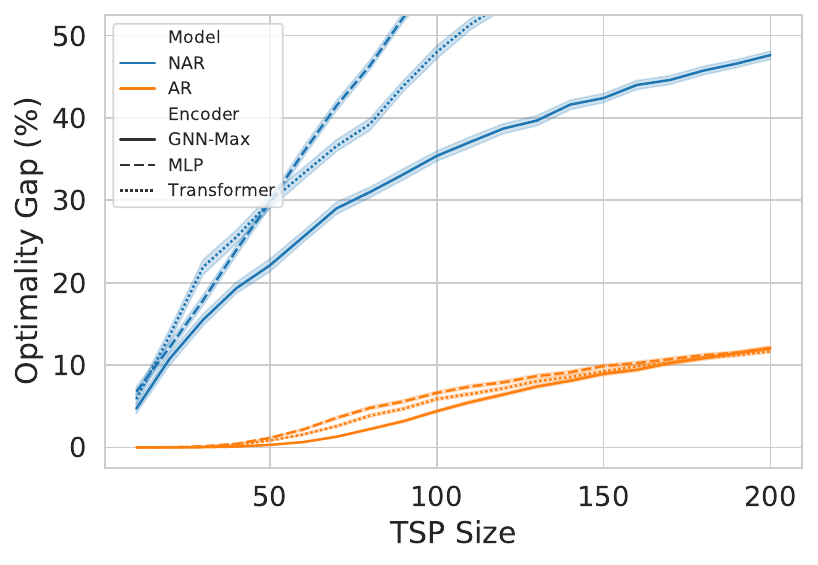}
    \caption{\small\textbf{Comparing AR and NAR decoders.} Sequential AR decoding is a powerful inductive bias for TSP as it enables significantly better generalization, even in the absence of graph structure (MLP encoders).}
    \label{fig:decoder}
\end{minipage}
\quad
\begin{minipage}{.47\textwidth}
\centering
    \includegraphics[width=0.98\linewidth]{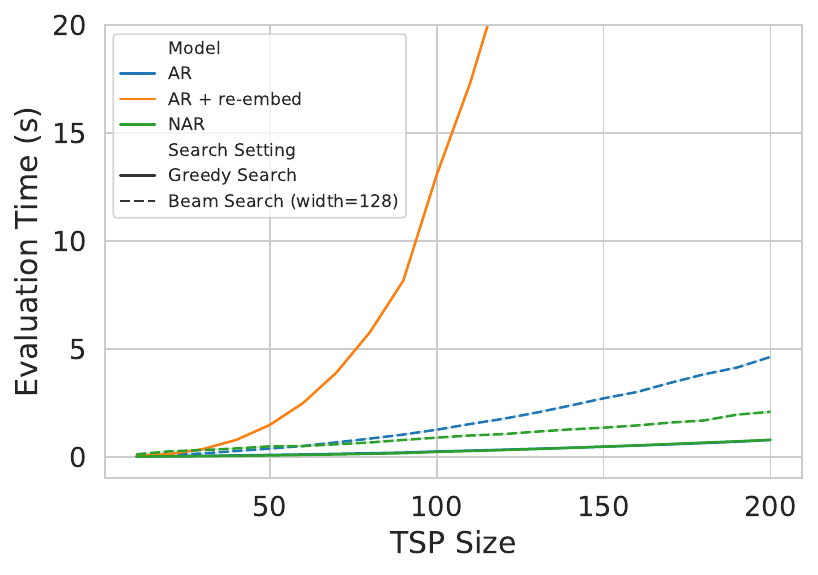}
    \caption{\small\textbf{Inference time for various decoders.} One-shot NAR decoding is significantly faster than sequential AR, especially when re-embedding the graph at each decoding step~\cite{khalil2017learning}.}
    \label{fig:decoder_timing}
\end{minipage}
\end{figure}


\subsection{Which decoder has a better inductive bias for TSP?}

Figure~\ref{fig:decoder} compares NAR and AR decoders for identical models. 
To isolate the impact of the decoder's inductive bias without the inductive bias imposed by GNNs, we also show Transformer encoders on full graphs as well as structure-agnostic MLPs.
Within our experimental setup, AR decoders are able to fit the training data as well as generalize significantly better than NAR decoders,
indicating that sequential decoding is powerful for TSP even without graph information.

Conversely, NAR architectures are a poor inductive bias as they require significantly more computation to perform competitively to AR decoders.
For instance, recent models~\cite{nowak2017note,joshi2019efficient} used more than 30 GNN layers with over 10~Million parameters.
We believe that such overparameterized networks are able to memorize all patterns for small TSP training sizes~\cite{zhang2016understanding},
but the learnt policy is unable to generalize beyond training graph sizes.
At the same time, when compared fairly within the same experimental settings, NAR decoders are significantly faster than AR decoders described in Section~\ref{sec:method:decoder} as well as those which re-embed the graph at each decoding step~\cite{khalil2017learning}, see Figure~\ref{fig:decoder_timing}.


\subsection{How do learning paradigms impact the search phase?}
\label{sec:experiments:search}

Identical models are trained via supervised learning (SL) and reinforcement learning (RL)\footnote{For RL, we show the greedy rollout baseline. Critic baseline results are available in Appendix~\ref{app:omitted}}.
Figure~\ref{fig:learning_and_search} illustrates that, when using greedy decoding during inference, RL models perform better on the training size as well as on larger graphs. 
Conversely, SL models improve over their RL counterparts when performing beam search or sampling.

In Appendix~\ref{app:learning-and-search}, we find that the rollout baseline, which encourages better greedy behaviour, leads to the model making very confident predictions about selecting the next node at each decoding step, even out of training size range.
In contrast, SL models are trained with teacher forcing, \textit{i.e.} imitating the optimal solver at each step instead of using their own prediction. 
This results in less confident predictions and poor greedy decoding, but makes the probability distribution more amenable to beam search and sampling, as shown in Figure~\ref{fig:bs_and_learning}.
Our results advocate for tighter coupling between the training and inference phase of learning-driven TSP solvers, mirroring recent findings in generative models for text~\cite{Holtzman2020The}.



\begin{figure}[t!]
\centering
\begin{minipage}{.48\textwidth}
\centering
    \includegraphics[width=0.98\linewidth]{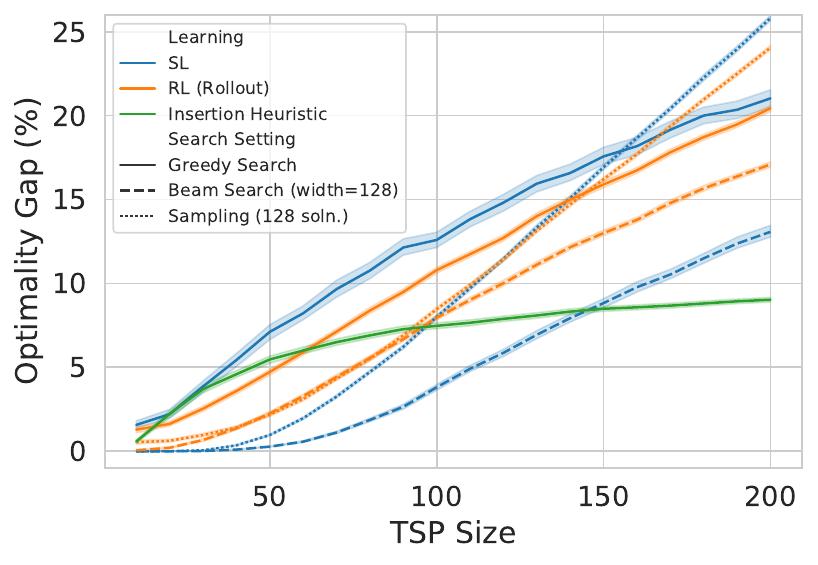}
    \caption{\small\textbf{Comparing solution search settings.} Under greedy decoding, RL demonstrates better performance and generalization.  Conversely, SL models improve over their RL counterparts when performing beam search or sampling.}
    \label{fig:learning_and_search}
\end{minipage}
\quad
\begin{minipage}{.48\textwidth}
\centering
    \includegraphics[width=0.98\linewidth]{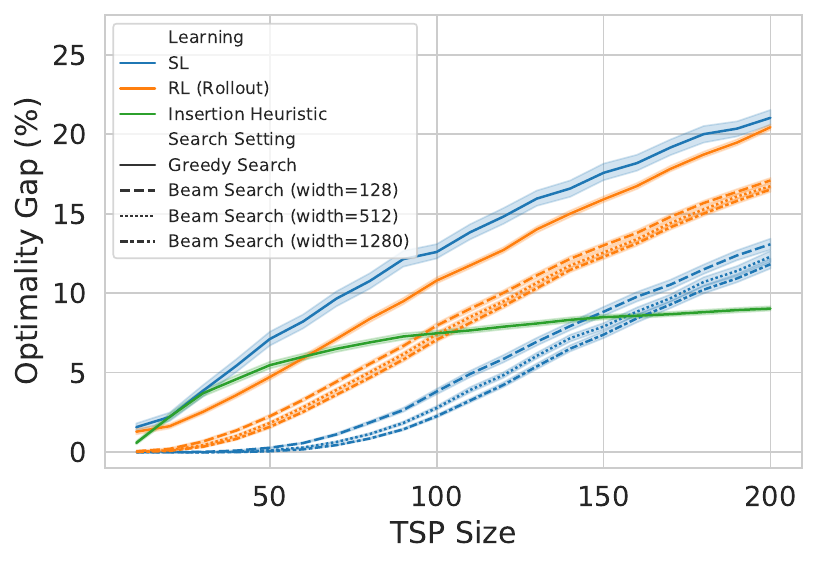}
    \caption{\small\textbf{Impact of increasing beam width.} Teacher-forcing during SL leads to poor generalization under greedy decoding, but makes the probability distribution more amenable to beam search.}
    \label{fig:bs_and_learning}
\end{minipage}
\end{figure}


\subsection{Which learning paradigm scales better?}
\label{sec:experiments:scale}

Our experiments till this point have focused on isolating the impact of various pipeline components on zero-shot generalization under limited computation.
At the same time, recent results in natural language processing have highlighted the power of large scale pre-training for effective transfer learning~\cite{raffel2019exploring}. 
To better understand the impact of learning paradigms when scaling computation, we double the model parameters (up to 750,000) and train on tens times more data (12.8M samples) for AR architectures.
We monitor optimality gap on the training size range~(TSP20-50) as well as a larger size~(TSP100) vs. the number of training samples.

\begin{figure}[t!]
    \centering
    \includegraphics[width=0.55\linewidth]{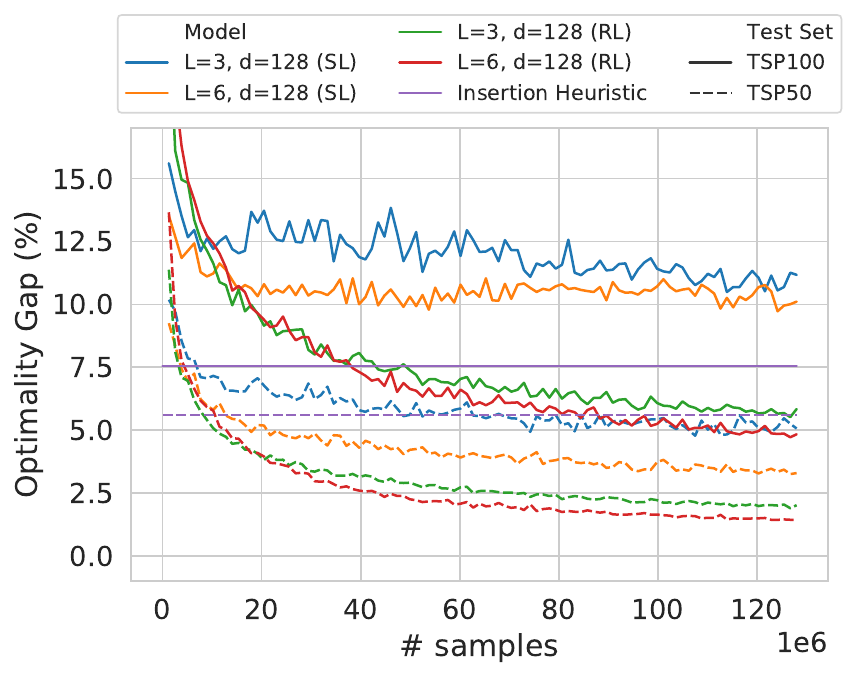}
    \caption{\small\textbf{Scaling computation and parameters for SL and RL-trained models.} All models are trained on TSP20-50. 
    We plot optimality gap on 1,280 held-out samples of both TSP50 (performance on training size) and TSP100 (out-of-distribution generalization) under greedy decoding.
    Note that SL models are less amenable than RL models to greedy search.
    RL models are able to keep improving their performance within as well as outside of training size range with more data.
    On the other hand, SL performance is bottlenecked by the need for optimal groundtruth solutions.
    }
    \label{fig:scale-ar}
\end{figure}

In Figure~\ref{fig:scale-ar}, we see that increasing model capacity leads to better learning.
Notably, RL models, which train on unique randomly generated samples throughout, are able to keep improving their performance within as well as outside of training size range as they see more samples.
On the other hand, SL is bottlenecked by the need for optimal groundtruth solutions: SL models iterate over the same 1.28M unique labelled samples and stop improving at a point.
Beyond favorable inductive biases, distributed and sample-efficient RL algorithms~\cite{schulman2017proximal} may be a key ingredient for learning from and scaling up to larger TSPs beyond tens of nodes.




\section{Recent Case Studies and Future Work}
Since the initial publication of this work \cite{joshi2021learning}, deep learning for routing problems has received considerable attention from the research community \cite{wu2019learning, kwon2020pomo, fu2020generalize, pmlr-v129-costa20a, kool2021deep, xin2021neurolkh, ma2021learning, ouyang2021generalization, hudson2021graph}.
In this section and in an associated blogpost~\cite{chaitanyak2022recentadvancesin}, we highlight recent advances, characterize them using the unified pipeline presented in Figure~\ref{fig:pipeline}, and provide future research directions, with a focus on improving generalization to large-scale and real-world instances.
As a reminder, the unified neural combinatorial optimization pipeline consists of:
(1) Problem Definition $\rightarrow$
(2) Graph Embedding $\rightarrow$
(3) Solution Decoding $\rightarrow$
(4) Solution Search $\rightarrow$
(5) Policy Learning.

\textbf{Leveraging equivariance and symmetries\quad}
The autoregressive Attention Model \cite{kool2018attention} sequentially constructs TSP tours as permutations of cities, but does not consider the underlying symmetries of routing problems. 

Kwon et al. \cite{kwon2020pomo} consider invariance to the starting city in constructive heuristics:
They propose to train the Attention Model with a new reinforcement learning algorithm (innovating on box 5 in Figure~\ref{fig:pipeline}(a)) which exploits the existence of multiple optimal tour permutations.
Similarly, Ouyang, Wang, et al. \cite{ouyang2021generalization} consider invariance with respect to rotations, reflections, and translations (Euclidean symmetry group) of the input cities:
They propose a constructive approach similar to Attention Model while ensuring invariance by performing data augmentation during the problem definition stage (Figure~\ref{fig:pipeline}(a), box 1) and using relative coordinates during graph encoding (Figure~\ref{fig:pipeline}(a), box 2).
Their approach shows particularly strong results on zero-shot generalization from random instances to the real-world TSPLib benchmark suite.

Future work may follow the Geometric Deep Learning blueprint \cite{bronstein2021geometric} by designing models that respect the symmetries and inductive biases that govern the data.
As routing problems are embedded in euclidean coordinates and the routes are cyclical, incorporating these contraints directly into the architectures or learning paradigms may be a principled approach to improving generalization to large-scale instances greater than those seen during training.

\textbf{Improved graph search algorithms\quad}
Several papers have proposed to improve the one-shot non-autoregressive approach of Joshi et al. \cite{joshi2019efficient} by retaining the same GNN encoder (Figure~\ref{fig:pipeline}(a), box 2) while replacing the graph search component of the pipeline (Figure~\ref{fig:pipeline}(a), box 4) with more powerful and flexible algorithms, \textit{e.g.} Dynamic Programming \cite{kool2021deep} or Monte-Carlo Tree Search (MCTS) \cite{fu2020generalize}.

Notably, the GNN + MCTS framework of Fu et al. \cite{fu2020generalize} shows that the NAR approach can generalize to TSPs with up to 1000 nodes.
They ensure that the predictions of the GNN encoder generalize from small to large TSP by updating the problem definition (Figure~\ref{fig:pipeline}(a), box 1): 
large problem instances are represented as many smaller sub-graphs which are of the same size as the training graphs for the GNN, and then merge the GNN edge predictions before performing MCTS.

Overall, this line of work suggests that stronger coupling between the design of both the neural and symbolic/search components of models is essential for out-of-distribution generalization.

\textbf{Learning within local search heuristics\quad}
Recent work has explored an alternative to constructive AR and NAR decoding schemes which involves learning to iteratively improve (sub-optimal) solutions or learning to perform local search \cite{wu2019learning, pmlr-v129-costa20a, xin2021neurolkh, ma2021learning, hudson2021graph}.
Since deep learning is used to guide decisions within classical search algorithms (which are designed to work regardless of problem scale), this approach implicitly leads to better zero-shot generalization to larger problem instances compared to constructive approaches studied in our work.
In particular, NeuroLKH \cite{xin2021neurolkh} uses GNNs to improve the Lin-Kernighan-Helsgaun algorithm and demonstrates strong zero-shot generalization to TSP with 5000 nodes as well as across TSPLib instances. 

A limitation of this line of work is the need for hand-designed local search heuristics, which may be missing for understudied problems.
On the other hand, constructive approaches are comparatively easier to adapt to new problems by enforcing constraints during the solution decoding and search procedure (Figure~\ref{fig:pipeline}(a), box 4).

\textbf{Learning Paradigms that promote generalization\quad}
Future work could look at novel learning paradigms which explicitly focus on generalization beyond supervised and reinforcement learning.
For \textit{e.g.}, this work explored zero-shot generalization to larger problems, but the logical next step is to fine-tune the model on a small number of larger problem instances \cite{hottung2021efficient}. 
Thus, it will be interesting to explore fine-tuning/generalization as a meta-learning problem, wherein the goal is to train model parameters specifically for fast adaptation and fine-tuning to new data distributions and problem sizes.

Another interesting direction could explore tackling understudied routing problems with challenging constraints via multi-task pre-training on well-known routing problems such as TSP and CVPR, followed by problem-specific finetuning.
Similar to language modelling as a pre-training objective in NLP \cite{raffel2019exploring}, the goal of pre-training for routing would be to learn generally useful neural network representations that can transfer well to novel routing problems. 


\section{Conclusion}
\label{sec:conclusion}

Learning-driven solvers for combinatorial problems such as the Travelling Salesperson Problem have shown promising results for trivially small instances up to a few hundred nodes. 
However, scaling fully \textit{end-to-end} deep learning approaches to real-world instances is still an open question as training on large graphs is extremely time-consuming and challenging to learn from.

This paper advocates for an alternative to expensive large-scale training: 
training models efficiently on trivially small TSP and transferring the learnt policy to larger graphs in a \textit{zero-shot} fashion or via fast fine-tuning. 
Thus, identifying promising inductive biases, architectures and learning paradigms that enable such zero-shot generalization to large and more complex instances is a key concern for tackling real-world combinatorial problems.

We perform the first principled investigation into zero-shot generalization for learning large scale TSP, 
unifying state-of-the-art architectures and learning paradigms into one experimental pipeline for neural combinatorial optimization.
Our findings suggest that key design choices such as GNN layers, normalization schemes, graph sparsification, and learning paradigms need to be explicitly re-designed to consider out-of-distribution generalization. 
Additionally, we use our unified pipeline to characterize recent advances in deep learning for routing problems and provide new directions to stimulate future research.


\section*{Acknowledgements}
We would like to thank R. Anand, X. Bresson, V. Dwivedi, A. Ferber, E. Khalil, W. Kool, R. Levie, A. Prouvost, P. Veličković and the anonymous reviewers for helpful comments and discussions.


\bibliographystyle{abbrv}
\bibliography{references}

\begin{thebibliography}{10}

\bibitem{abadi2016tensorflow}
M.~Abadi, A.~Agarwal, P.~Barham, E.~Brevdo, Z.~Chen, C.~Citro, G.~S. Corrado,
  A.~Davis, J.~Dean, M.~Devin, et~al.
\newblock Tensorflow: Large-scale machine learning on heterogeneous distributed
  systems.
\newblock {\em arXiv preprint}, 2016.

\bibitem{applegate2006concorde}
D.~Applegate, R.~Bixby, V.~Chvatal, and W.~Cook.
\newblock Concorde tsp solver, 2006.

\bibitem{applegate2006traveling}
D.~L. Applegate, R.~E. Bixby, V.~Chvatal, and W.~J. Cook.
\newblock {\em The traveling salesman problem: a computational study}.
\newblock 2006.

\bibitem{ba2016layer}
J.~L. Ba, J.~R. Kiros, and G.~E. Hinton.
\newblock Layer normalization.
\newblock {\em arXiv preprint}, 2016.

\bibitem{battaglia2018relational}
P.~W. Battaglia, J.~B. Hamrick, V.~Bapst, A.~Sanchez-Gonzalez, V.~Zambaldi,
  M.~Malinowski, A.~Tacchetti, D.~Raposo, A.~Santoro, R.~Faulkner, et~al.
\newblock Relational inductive biases, deep learning, and graph networks.
\newblock {\em arXiv preprint}, 2018.

\bibitem{bello2016neural}
I.~Bello, H.~Pham, Q.~V. Le, M.~Norouzi, and S.~Bengio.
\newblock Neural combinatorial optimization with reinforcement learning.
\newblock In {\em ICLR}, 2017.

\bibitem{bengio2018machine}
Y.~Bengio, A.~Lodi, and A.~Prouvost.
\newblock Machine learning for combinatorial optimization: a methodological
  tour d'horizon.
\newblock {\em European Journal of Operational Research}, 2020.

\bibitem{bresson2018experimental}
X.~Bresson and T.~Laurent.
\newblock An experimental study of neural networks for variable graphs.
\newblock In {\em ICLR Workshop}, 2018.

\bibitem{bresson2019two}
X.~Bresson and T.~Laurent.
\newblock A two-step graph convolutional decoder for molecule generation.
\newblock In {\em NeurIPS Workshop on Machine Learning and the Physical
  Sciences}, 2019.

\bibitem{bronstein2021geometric}
M.~M. Bronstein, J.~Bruna, T.~Cohen, and P.~Veličković.
\newblock Geometric deep learning: Grids, groups, graphs, geodesics, and
  gauges.
\newblock {\em arXiv preprint}, 2021.

\bibitem{cappart2021combinatorial}
Q.~Cappart, D.~Ch{\'e}telat, E.~Khalil, A.~Lodi, C.~Morris, and
  P.~Veli{\v{c}}kovi{\'c}.
\newblock Combinatorial optimization and reasoning with graph neural networks.
\newblock In {\em IJCAI}, 2021.

\bibitem{cappart2019improving}
Q.~Cappart, E.~Goutierre, D.~Bergman, and L.-M. Rousseau.
\newblock Improving optimization bounds using machine learning: Decision
  diagrams meet deep reinforcement learning.
\newblock In {\em AAAI}, 2019.

\bibitem{chalumeau2021seapearl}
F.~Chalumeau, I.~Coulon, Q.~Cappart, and L.-M. Rousseau.
\newblock Seapearl: A constraint programming solver guided by reinforcement
  learning.
\newblock In {\em CPAIOR}, 2021.

\bibitem{chen2019learning}
X.~Chen and Y.~Tian.
\newblock Learning to perform local rewriting for combinatorial optimization.
\newblock In {\em NeurIPS}, 2019.

\bibitem{corso2020principal}
G.~Corso, L.~Cavalleri, D.~Beaini, P.~Li{\`o}, and P.~Veli{\v{c}}kovi{\'c}.
\newblock Principal neighbourhood aggregation for graph nets.
\newblock In {\em NeurIPS}, 2020.

\bibitem{pmlr-v129-costa20a}
P.~R. d.~O. da~Costa, J.~Rhuggenaath, Y.~Zhang, and A.~Akcay.
\newblock Learning 2-opt heuristics for the traveling salesman problem via deep
  reinforcement learning.
\newblock In {\em Asian Conference on Machine Learning}, 2020.

\bibitem{deudon2018learning}
M.~Deudon, P.~Cournut, A.~Lacoste, Y.~Adulyasak, and L.-M. Rousseau.
\newblock Learning heuristics for the tsp by policy gradient.
\newblock In {\em CPAIOR}, 2018.

\bibitem{dwivedi2020benchmarking}
V.~P. Dwivedi, C.~K. Joshi, T.~Laurent, Y.~Bengio, and X.~Bresson.
\newblock Benchmarking graph neural networks.
\newblock {\em arXiv preprint}, 2020.

\bibitem{ferber2020mipaal}
A.~Ferber, B.~Wilder, B.~Dilkina, and M.~Tambe.
\newblock Mipaal: Mixed integer program as a layer.
\newblock In {\em AAAI}, 2020.

\bibitem{francois2019how}
A.~François, Q.~Cappart, and L.-M. Rousseau.
\newblock How to evaluate machine learning approaches for combinatorial
  optimization: Application to the travelling salesman problem.
\newblock {\em arXiv preprint}, 2019.

\bibitem{fu2020generalize}
Z.-H. Fu, K.-B. Qiu, and H.~Zha.
\newblock Generalize a small pre-trained model to arbitrarily large tsp
  instances.
\newblock In {\em AAAI}, 2021.

\bibitem{garg2020generalization}
V.~K. Garg, S.~Jegelka, and T.~Jaakkola.
\newblock Generalization and representational limits of graph neural networks.
\newblock In {\em ICML}, 2020.

\bibitem{gasse2019exact}
M.~Gasse, D.~Ch{\'e}telat, N.~Ferroni, L.~Charlin, and A.~Lodi.
\newblock Exact combinatorial optimization with graph convolutional neural
  networks.
\newblock In {\em NeurIPS}, 2019.

\bibitem{gilmer2017neural}
J.~Gilmer, S.~S. Schoenholz, P.~F. Riley, O.~Vinyals, and G.~E. Dahl.
\newblock Neural message passing for quantum chemistry.
\newblock In {\em ICML}, 2017.

\bibitem{gomez2018automatic}
R.~G{\'o}mez-Bombarelli, J.~N. Wei, D.~Duvenaud, J.~M. Hern{\'a}ndez-Lobato,
  B.~S{\'a}nchez-Lengeling, D.~Sheberla, J.~Aguilera-Iparraguirre, T.~D.
  Hirzel, R.~P. Adams, and A.~Aspuru-Guzik.
\newblock Automatic chemical design using a data-driven continuous
  representation of molecules.
\newblock {\em ACS central science}, 2018.

\bibitem{hermans2017defense}
A.~Hermans, L.~Beyer, and B.~Leibe.
\newblock In defense of the triplet loss for person re-identification.
\newblock {\em arXiv preprint}, 2017.

\bibitem{Holtzman2020The}
A.~Holtzman, J.~Buys, L.~Du, M.~Forbes, and Y.~Choi.
\newblock The curious case of neural text degeneration.
\newblock In {\em ICLR}, 2020.

\bibitem{hottung2021efficient}
A.~Hottung, Y.-D. Kwon, and K.~Tierney.
\newblock Efficient active search for combinatorial optimization problems.
\newblock {\em arXiv preprint}, 2021.

\bibitem{huang2019coloring}
J.~Huang, M.~Patwary, and G.~Diamos.
\newblock Coloring big graphs with alphagozero.
\newblock {\em arXiv preprint}, 2019.

\bibitem{hudson2021graph}
B.~Hudson, Q.~Li, M.~Malencia, and A.~Prorok.
\newblock Graph neural network guided local search for the traveling
  salesperson problem.
\newblock {\em arXiv preprint}, 2021.

\bibitem{gurobi2015gurobi}
G.~O. Inc.
\newblock Gurobi optimizer reference manual.
\newblock {\em URL \url{http://www.gurobi.com}}, 2015.

\bibitem{ioffe2015batch}
S.~Ioffe and C.~Szegedy.
\newblock Batch normalization: Accelerating deep network training by reducing
  internal covariate shift.
\newblock {\em arXiv preprint}, 2015.

\bibitem{jin2018junction}
W.~Jin, R.~Barzilay, and T.~Jaakkola.
\newblock Junction tree variational autoencoder for molecular graph generation.
\newblock In {\em ICML}, 2018.

\bibitem{joshi2020transformers}
C.~Joshi.
\newblock Transformers are graph neural networks.
\newblock {\em The Gradient}, 2020.

\bibitem{chaitanyak2022recentadvancesin}
C.~K. Joshi and R.~Anand.
\newblock Recent advances in deep learning for routing problems.
\newblock In {\em ICLR Blog Track}, 2022.

\bibitem{joshi2021learning}
C.~K. Joshi, Q.~Cappart, L.-M. Rousseau, and T.~Laurent.
\newblock Learning tsp requires rethinking generalization.
\newblock In {\em International Conference on Principles and Practice of
  Constraint Programming}, 2021.

\bibitem{joshi2019efficient}
C.~K. Joshi, T.~Laurent, and X.~Bresson.
\newblock An efficient graph convolutional network technique for the travelling
  salesman problem.
\newblock {\em arXiv preprint}, 2019.

\bibitem{joshi2019learning}
C.~K. Joshi, T.~Laurent, and X.~Bresson.
\newblock On learning paradigms for the travelling salesman problem.
\newblock {\em NeurIPS Graph Representation Learning Workshop}, 2019.

\bibitem{khalil2017learning}
E.~Khalil, H.~Dai, Y.~Zhang, B.~Dilkina, and L.~Song.
\newblock Learning combinatorial optimization algorithms over graphs.
\newblock In {\em NeurIPS}, 2017.

\bibitem{kipf2017semi}
T.~N. Kipf and M.~Welling.
\newblock Semi-supervised classification with graph convolutional networks.
\newblock In {\em ICLR}, 2017.

\bibitem{kool2021deep}
W.~Kool, H.~van Hoof, J.~Gromicho, and M.~Welling.
\newblock Deep policy dynamic programming for vehicle routing problems.
\newblock {\em arXiv preprint}, 2021.

\bibitem{kool2018attention}
W.~Kool, H.~van Hoof, and M.~Welling.
\newblock Attention, learn to solve routing problems!
\newblock In {\em ICLR}, 2019.

\bibitem{kwon2020pomo}
Y.-D. Kwon, J.~Choo, B.~Kim, I.~Yoon, Y.~Gwon, and S.~Min.
\newblock Pomo: Policy optimization with multiple optima for reinforcement
  learning.
\newblock In {\em NeurIPS}, 2020.

\bibitem{lenstra1975some}
J.~K. Lenstra and A.~R. Kan.
\newblock Some simple applications of the travelling salesman problem.
\newblock {\em Journal of the Operational Research Society}, 1975.

\bibitem{levie2019transferability}
R.~Levie, M.~M. Bronstein, and G.~Kutyniok.
\newblock Transferability of spectral graph convolutional neural networks.
\newblock {\em arXiv preprint}, 2019.

\bibitem{li2018combinatorial}
Z.~Li, Q.~Chen, and V.~Koltun.
\newblock Combinatorial optimization with graph convolutional networks and
  guided tree search.
\newblock In {\em NeurIPS}, 2018.

\bibitem{ma2019combinatorial}
Q.~Ma, S.~Ge, D.~He, D.~Thaker, and I.~Drori.
\newblock Combinatorial optimization by graph pointer networks and hierarchical
  reinforcement learning.
\newblock In {\em AAAI Workshop on Deep Learning on Graphs}, 2020.

\bibitem{ma2021learning}
Y.~Ma, J.~Li, Z.~Cao, W.~Song, L.~Zhang, Z.~Chen, and J.~Tang.
\newblock Learning to iteratively solve routing problems with dual-aspect
  collaborative transformer.
\newblock In {\em NeurIPS}, 2021.

\bibitem{mao2019learning}
H.~Mao, M.~Schwarzkopf, S.~B. Venkatakrishnan, Z.~Meng, and M.~Alizadeh.
\newblock Learning scheduling algorithms for data processing clusters.
\newblock In {\em ACM Special Interest Group on Data Communication}, 2019.

\bibitem{mirhoseini2021graph}
A.~Mirhoseini, A.~Goldie, M.~Yazgan, J.~W. Jiang, E.~Songhori, S.~Wang, Y.-J.
  Lee, E.~Johnson, O.~Pathak, A.~Nazi, et~al.
\newblock A graph placement methodology for fast chip design.
\newblock {\em Nature}, 2021.

\bibitem{mirhoseini2017device}
A.~Mirhoseini, H.~Pham, Q.~V. Le, B.~Steiner, R.~Larsen, Y.~Zhou, N.~Kumar,
  M.~Norouzi, S.~Bengio, and J.~Dean.
\newblock Device placement optimization with reinforcement learning.
\newblock In {\em ICML}, 2017.

\bibitem{nazari2018reinforcement}
M.~Nazari, A.~Oroojlooy, L.~Snyder, and M.~Tak{\'a}c.
\newblock Reinforcement learning for solving the vehicle routing problem.
\newblock In {\em NeurIPS}, 2018.

\bibitem{nowak2018divide}
A.~Nowak, D.~Folqu{\'e}, and J.~B. Estrach.
\newblock Divide and conquer networks.
\newblock In {\em ICLR}, 2018.

\bibitem{nowak2017note}
A.~Nowak, S.~Villar, A.~S. Bandeira, and J.~Bruna.
\newblock A note on learning algorithms for quadratic assignment with graph
  neural networks.
\newblock {\em arXiv preprint}, 2017.

\bibitem{ouyang2021generalization}
W.~Ouyang, Y.~Wang, P.~Weng, and S.~Han.
\newblock Generalization in deep rl for tsp problems via equivariance and local
  search.
\newblock {\em arXiv preprint}, 2021.

\bibitem{paliwal2019regal}
A.~Paliwal, F.~Gimeno, V.~Nair, Y.~Li, M.~Lubin, P.~Kohli, and O.~Vinyals.
\newblock Regal: Transfer learning for fast optimization of computation graphs.
\newblock {\em arXiv preprint}, 2019.

\bibitem{raffel2019exploring}
C.~Raffel, N.~Shazeer, A.~Roberts, K.~Lee, S.~Narang, M.~Matena, Y.~Zhou,
  W.~Li, and P.~J. Liu.
\newblock Exploring the limits of transfer learning with a unified text-to-text
  transformer.
\newblock {\em JMLR}, 2020.

\bibitem{sato2019approximation}
R.~Sato, M.~Yamada, and H.~Kashima.
\newblock Approximation ratios of graph neural networks for combinatorial
  problems.
\newblock In {\em NeurIPS}, 2019.

\bibitem{schulman2017proximal}
J.~Schulman, F.~Wolski, P.~Dhariwal, A.~Radford, and O.~Klimov.
\newblock Proximal policy optimization algorithms.
\newblock {\em arXiv preprint}, 2017.

\bibitem{selsam2018learning}
D.~Selsam, M.~Lamm, B.~B{\"u}nz, P.~Liang, L.~de~Moura, and D.~L. Dill.
\newblock Learning a sat solver from single-bit supervision.
\newblock In {\em ICLR}, 2019.

\bibitem{senior2020improved}
A.~W. Senior, R.~Evans, J.~Jumper, J.~Kirkpatrick, L.~Sifre, T.~Green, C.~Qin,
  A.~{\v{Z}}{\'\i}dek, A.~W. Nelson, A.~Bridgland, et~al.
\newblock Improved protein structure prediction using potentials from deep
  learning.
\newblock {\em Nature}, 2020.

\bibitem{sutskever2014sequence}
I.~Sutskever, O.~Vinyals, and Q.~V. Le.
\newblock Sequence to sequence learning with neural networks.
\newblock In {\em NeurIPS}, 2014.

\bibitem{vaswani2017attention}
A.~Vaswani, N.~Shazeer, N.~Parmar, J.~Uszkoreit, L.~Jones, A.~N. Gomez,
  {\L}.~Kaiser, and I.~Polosukhin.
\newblock Attention is all you need.
\newblock In {\em NeurIPS}, 2017.

\bibitem{velivckovic2021neural}
P.~Veli{\v{c}}kovi{\'c} and C.~Blundell.
\newblock Neural algorithmic reasoning.
\newblock {\em Patterns}, 2021.

\bibitem{velickovic2018graph}
P.~Veli{\v{c}}kovi{\'{c}}, G.~Cucurull, A.~Casanova, A.~Romero, P.~Li{\`{o}},
  and Y.~Bengio.
\newblock {Graph Attention Networks}.
\newblock {\em ICLR}, 2018.

\bibitem{velickovic2019neural}
P.~Veličković, R.~Ying, M.~Padovano, R.~Hadsell, and C.~Blundell.
\newblock Neural execution of graph algorithms.
\newblock In {\em ICLR}, 2020.

\bibitem{vinyals2015pointer}
O.~Vinyals, M.~Fortunato, and N.~Jaitly.
\newblock Pointer networks.
\newblock In {\em NeurIPS}, 2015.

\bibitem{wilder2019melding}
B.~Wilder, B.~Dilkina, and M.~Tambe.
\newblock Melding the data-decisions pipeline: Decision-focused learning for
  combinatorial optimization.
\newblock In {\em AAAI}, 2019.

\bibitem{williams1992simple}
R.~J. Williams.
\newblock Simple statistical gradient-following algorithms for connectionist
  reinforcement learning.
\newblock {\em Machine learning}, 1992.

\bibitem{williams1989learning}
R.~J. Williams and D.~Zipser.
\newblock A learning algorithm for continually running fully recurrent neural
  networks.
\newblock {\em Neural computation}, 1(2):270--280, 1989.

\bibitem{wu2019learning}
Y.~Wu, W.~Song, Z.~Cao, J.~Zhang, and A.~Lim.
\newblock Learning improvement heuristics for solving routing problem.
\newblock {\em IEEE Transactions on Neural Networks and Learning Systems},
  2021.

\bibitem{xin2021neurolkh}
L.~Xin, W.~Song, Z.~Cao, and J.~Zhang.
\newblock Neurolkh: Combining deep learning model with lin-kernighan-helsgaun
  heuristic for solving the traveling salesman problem.
\newblock In {\em NeurIPS}, 2021.

\bibitem{xu2018powerful}
K.~Xu, W.~Hu, J.~Leskovec, and S.~Jegelka.
\newblock How powerful are graph neural networks?
\newblock In {\em ICLR}, 2019.

\bibitem{xu2019can}
K.~Xu, J.~Li, M.~Zhang, S.~S. Du, K.-i. Kawarabayashi, and S.~Jegelka.
\newblock What can neural networks reason about?
\newblock In {\em ICLR}, 2019.

\bibitem{xu2020neural}
K.~Xu, J.~Li, M.~Zhang, S.~S. Du, K.-i. Kawarabayashi, and S.~Jegelka.
\newblock How neural networks extrapolate: From feedforward to graph neural
  networks.
\newblock In {\em ICLR}, 2020.

\bibitem{ying2018hierarchical}
Z.~Ying, J.~You, C.~Morris, X.~Ren, W.~Hamilton, and J.~Leskovec.
\newblock Hierarchical graph representation learning with differentiable
  pooling.
\newblock In {\em NeurIPS}, 2018.

\bibitem{yolcu2019learning}
E.~Yolcu and B.~Poczos.
\newblock Learning local search heuristics for boolean satisfiability.
\newblock In {\em NeurIPS}, 2019.

\bibitem{you2018graph}
J.~You, B.~Liu, Z.~Ying, V.~Pande, and J.~Leskovec.
\newblock Graph convolutional policy network for goal-directed molecular graph
  generation.
\newblock In {\em NeurIPS}, 2018.

\bibitem{zhang2016understanding}
C.~Zhang, S.~Bengio, M.~Hardt, B.~Recht, and O.~Vinyals.
\newblock Understanding deep learning requires rethinking generalization.
\newblock In {\em ICLR}, 2017.

\bibitem{zhou2019gdp}
Y.~Zhou, S.~Roy, A.~Abdolrashidi, D.~Wong, P.~C. Ma, Q.~Xu, M.~Zhong, H.~Liu,
  A.~Goldie, A.~Mirhoseini, et~al.
\newblock Gdp: Generalized device placement for dataflow graphs.
\newblock {\em arXiv preprint}, 2019.

\end{thebibliography}

\appendix

\section{Additional Context for Figure 1}
\label{app:story}
\textbf{Experimental Setup\quad}
In Figure~\ref{fig:story}, we illustrate the computational challenges of learning large scale TSP by comparing three identical models trained on 12.8 Million TSP instances via reinforcement learning. 
Our experimental setup largely follows Section~\ref{sec:setup}.
All models use identical configurations: autoregressive decoding and Graph ConvNet encoder with \textsc{Max} aggregation and LayerNorm.
The TSP20-50 model is trained using the greedy rollout baseline~\cite{kool2018attention} and the Adam optimizer with batch size 128 and learning rate $1e-4$.
Direct training, active search and finetuning on TSP200 samples is done using learning rate $1e-5$, as we found larger learning rates to be unstable.
During active search and finetuning, we use an exponential moving average baseline, as recommended by Bello~et~al.~\cite{bello2016neural}.

\textbf{Furthest Insertion Baseline\quad}
We characterize `good' generalization across our experiments by the well-known \textit{furthest insertion} heuristic, 
which constructively builds a solution/partial tour $\pi'$ by inserting node $i$ between tour nodes $j_1, j_2 \in \pi'$ such that the distance from node $i$ to its nearest tour node $j_1$ is maximized.

We motivate our work by showing that learning from large TSP200 is intractable on university-scale hardware, and that efficient pre-training on trivial TSP20-50 enables models to better generalize to TSP200 in a zero-shot manner.
Within our computational budget, furthest insertion still outperforms our best models.
At the same time, we are not claiming that it is \textit{impossible} to outperform insertion heuristics with current approaches:
reinforcement learning-driven approaches will only continue to improve performance with more computation and training data.
We want to use simple non-learnt baselines to motivate the development of better architectures, learning paradigms and evaluation protocols for neural combinatorial optimization.

\textbf{Routing Problems and Generalization\quad}
It is worth mentioning why we chose to study TSP in particular. 
Firstly, TSP has stood the test of time in terms of relevance and continues to serve as an engine of discovery for general purpose techniques in applied mathematics.

TSP and associated routing problems have also emerged as a challenging testbed for learning-driven approaches to combinatorial optimization.
Whereas generalization to problem instances larger and more complex than those seen in training has at least partially been demonstrated on non-sequential problems such as SAT, MaxCut, MVC~\cite{khalil2017learning,li2018combinatorial,selsam2018learning}\footnote{
It is worth noting that classical algorithmic and symbolic components such as graph reduction, sophisticated tree search as well as post-hoc local search have been pivotal and complementary to GNNs in enabling such generalization.
},
the same architectures do not show strong generalization for TSP.
For \textit{e.g.}, furthest insertion outperforms or is competitive with state-of-the-art approaches for TSP above tens of nodes, see Figure~D.1.(e, f) from Khalil~et~al.~\cite{khalil2017learning} or Figure~5 from Kool~et~al.~\cite{kool2018attention}, despite using more computation and data than our controlled study.

\section{Hardware and Timings}
\label{app:hardware}
Fairly timing research code can be difficult due to differences in libraries used, hardware configurations and programmer skill.
In Table~\ref{tab:timing}, we report approximate total training time and inference time across TSP sizes for the model setup described in Section~\ref{sec:setup}.
All experiments were implemented in PyTorch and run on an Intel Xeon CPU E5-2690 v4 server and four Nvidia 1080Ti GPUs.
Four experiments were run on the server at any given time (each using a single GPU).
Training time may vary based on server load, thus we report the lowest training time across several runs in Table~\ref{tab:timing}.


\begin{table}[h!]
    \centering
    \caption{\small Approximate training time (12.8M samples) and inference time (1,280 samples) across TSP sizes and search settings for SL and RL-trained models. \textit{GS}: Greedy search, \textit{BS128}: beam search with width 128, \textit{S128}: sampling 128 solutions. RL training uses the rollout baseline and timing includes the time taken to update the baseline after each 128,000 samples.} 
    \label{tab:timing}
    \vspace{3mm}
    \begin{tabular}{cccccc}
        \toprule
        \multirow{2}{*}{\textbf{Graph Size}} & \multicolumn{2}{c}{\textbf{Training Time}} & \multicolumn{3}{c}{\textbf{Inference Time}} \\
         & \textbf{SL} & \textbf{RL} & \textbf{GS} & \textbf{BS128} & \textbf{S128} \\
        \midrule
        \midrule
        TSP20 & 4h 24m & 8h 02m & 2.62s & 7.06s & 63.37s \\
        TSP20-50 & 9h 49m & 15h 47m & - & - & - \\
        TSP50 & 16h 11m & 40h 29m & 7.45s & 29.09s & 86.48s \\
        TSP100 & 68h 34m & 108h 30m & 19.04s & 98.26s & 180.30s \\
        TSP200 & - & 495h 55m & 54.88s & 372.09s & 479.37s \\
        \bottomrule
    \end{tabular}
\end{table}


\section{Learning Paradigms and Amenity to Search}
\label{app:learning-and-search}

\begin{figure}[t!]
	\centering
	\subfloat[Supervised Learning]{
        \includegraphics[width=0.40\textwidth]{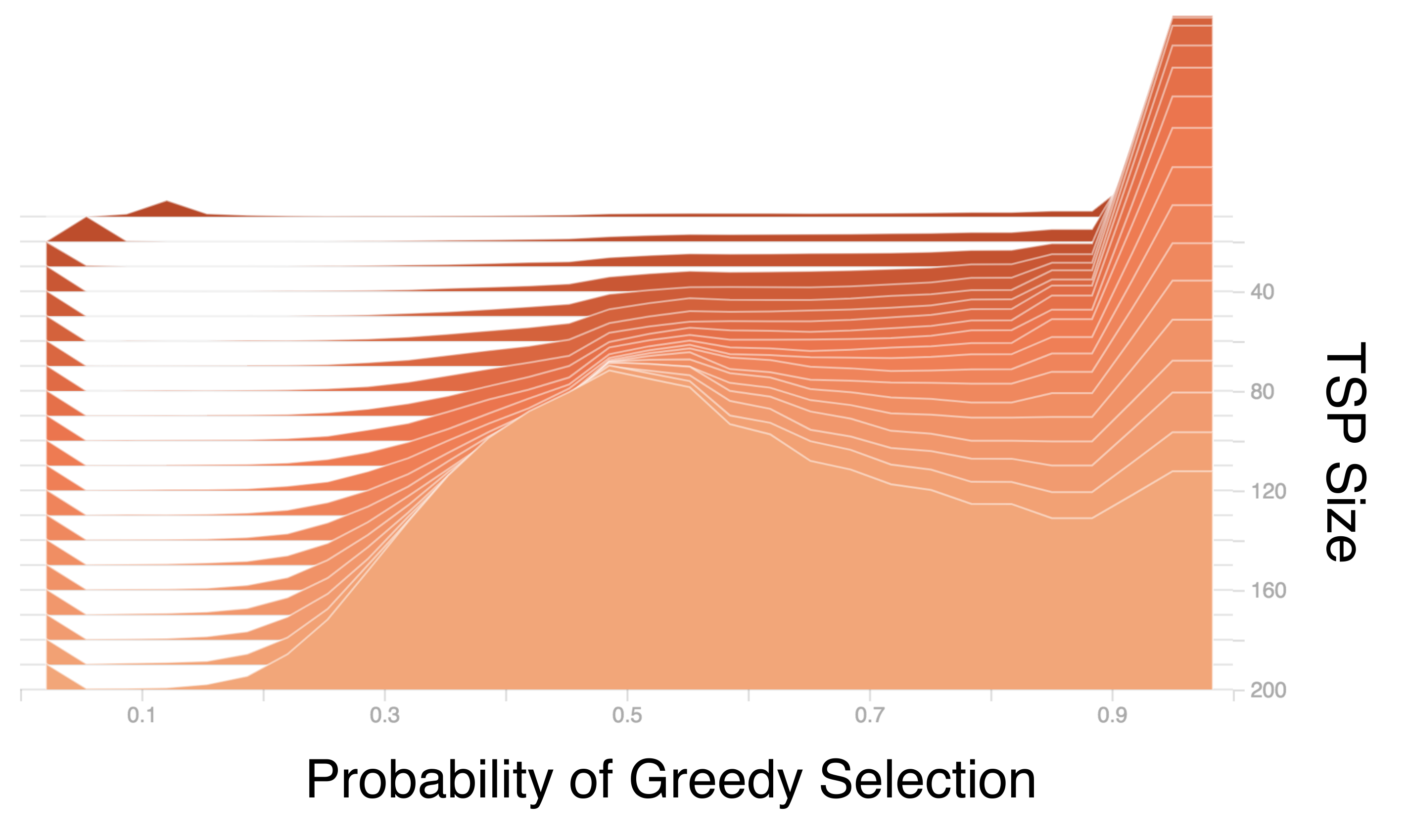}
        \label{fig:prob_selected_sl}
      }
    \quad
    \subfloat[Reinforcement Learning]{
        \includegraphics[width=0.45\textwidth]{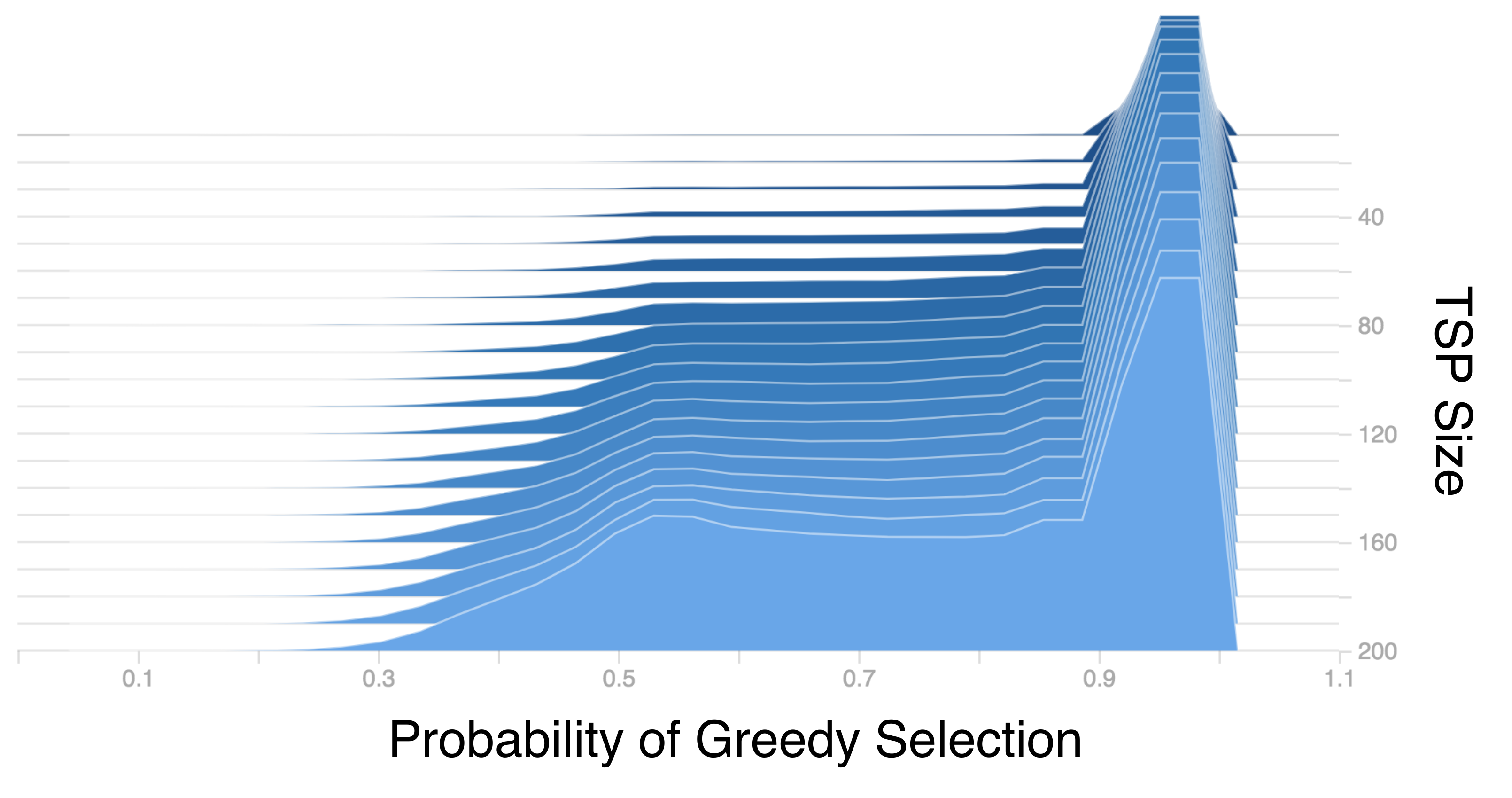}
        \label{fig:prob_selected_rl}
      }
    \caption{\small Histograms of greedy selection probabilities (x-axis) across TSP sizes (y-axis).
	}
	\label{fig:prob_selected}
\end{figure}

Figure~\ref{fig:bs_and_learning} demonstrate that SL models are more amenable to beam search and sampling, but are outperformed by RL-rollout models under greedy search.
In Figure~\ref{fig:prob_selected}, we investigate the impact of learning paradigms on probability distributions by plotting histograms of the probabilities of greedy selections during inference across TSP sizes for identical models trained with SL and RL.
We find that the rollout baseline, which encourages better greedy behaviour, leads to the model making very confident predictions about selecting the next node at each decoding step, even beyond training size range.
In contrast, SL models are trained with teacher forcing, \textit{i.e.} imitating the optimal solver at each step instead of using their own prediction. 
This results in less confident predictions and poor greedy decoding, but makes the probability distribution more amenable to beam search and sampling techniques.

We understand this phenomenon as follows:
More confident predictions (Figure~\ref{fig:prob_selected_rl}) do not automatically imply better solutions.
However, sampling repeatedly or maintaining the top-$b$ most probable solutions from such distributions is likely to contain very similar tours.
On the other hand, less sharp distributions (Figure~\ref{fig:prob_selected_sl}) are likely to yield more diverse tours with increasing $b$.
This may result in comparatively better optimality gap, especially for TSP sizes larger than those seen in training.


\section{Visualizing Node and Graph Embedding Spaces}
\label{app:graph-and-norm}
Our results in Section~\ref{sec:experiments:aggr-norm} suggest that inference beyond training sizes requires the development of GNN architectures and normalization layers that are both expressive as well as invariant to distribution shifts.
We explore how node and graph embeddings for TSP graphs evolve across training distribution (TSP20-50) and beyond (up to TSP200) through visualizing the statistics of the embedding spaces.
Intuitively, constructing TSP tours involves decisions which are not just locally optimal, but also optimal \textit{w.r.t} some global graph structure. 
Thus, node embeddings represent \textit{local} information while graph embeddings, which are conventionally computed as the mean of node embeddings, provide \textit{global} structural information. 

We utilize distribution plots
to study the variation in embedding statistics\footnote{
Distribution plots show 0, 5, 50, 95, and 100-percentiles for embedding statistics at various TSP sizes, thus visualizing how the statistics changes with problem scale (implemented via TensorBoard~\cite{abadi2016tensorflow}).}
of three identical models:
(1)~\textbf{\textit{GNN-Max}}, which represents our best model configuration from Section~\ref{sec:experiments}: autoregressive decoding, Graph ConvNet encoder with \textsc{Max} aggregation and BatchNorm with batch statistics;
(2)~\textbf{\textit{GNN-Sum}}, which uses \textsc{Sum} aggregation for the Graph ConvNet and shows comparatively poor generalization beyond training size, see Figure~\ref{fig:gnn_aggregation};
and (3)~\textbf{\textit{GNN-Max + learnt BN}}, which uses standard BatchNorm, \textit{i.e.} learns statistics from the training data, and also shows comparatively poor generalization, see Figure~\ref{fig:gnn_norm}.

We draw upon work in learning embeddings for computer vision~\cite{hermans2017defense} to characterize embedding spaces across TSP sizes according to:
(1)~\textbf{magnitudes}, denoted by $\ell_2$ norms, indicating whether embeddings are shrinking to one magnitude or expanding outwards as TSP size increases;
and (2)~\textbf{pair-wise distances}, which tells us how well-separated the embedding are, or whether they are pulled apart/towards each other as TSP size increases.


\begin{figure}[t!]
	\centering
	\subfloat[\textit{GNN-Sum}]{
        \includegraphics[width=0.3\textwidth]{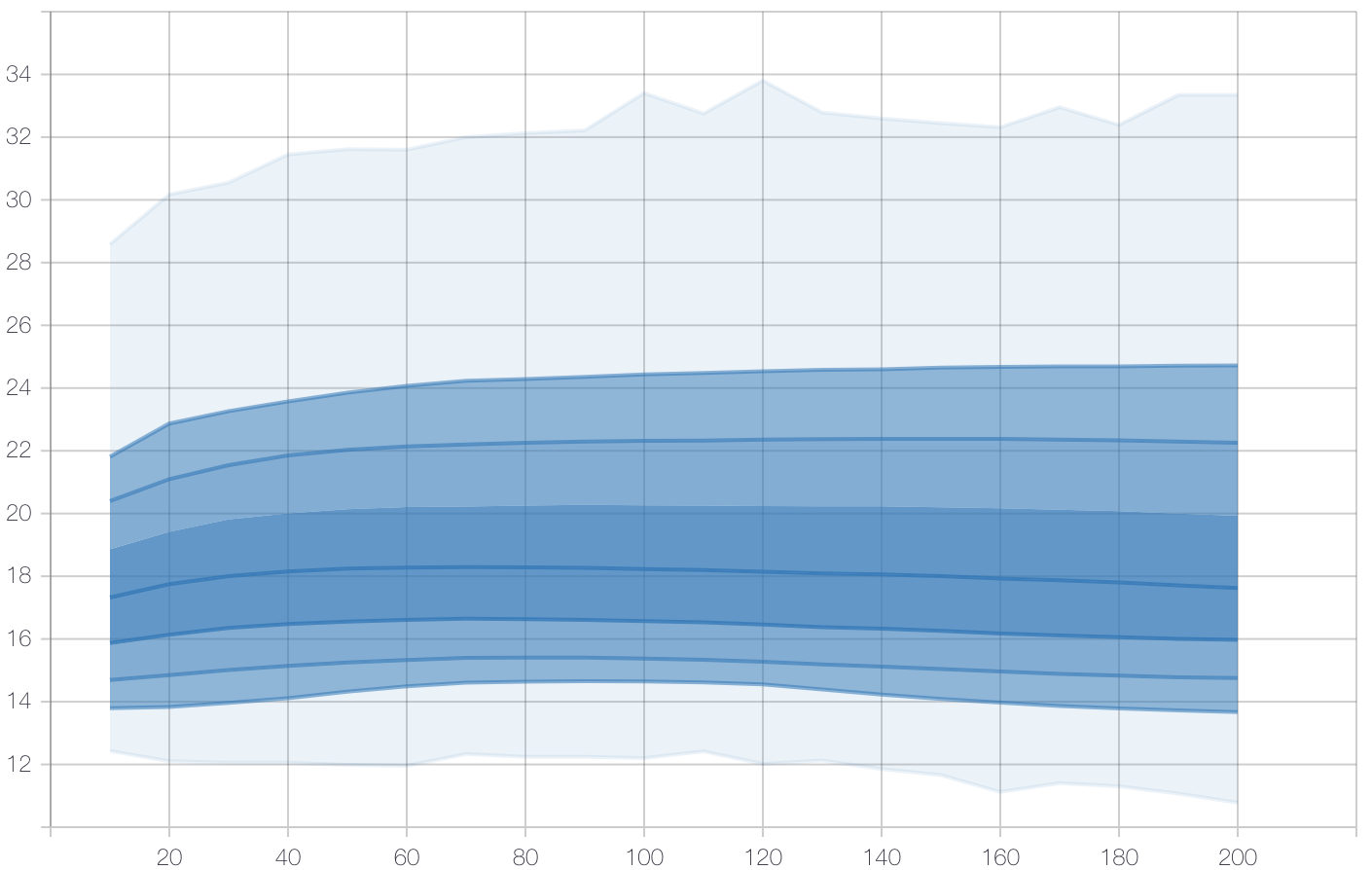}
        \label{fig:gnn-sum_emb_2norm}
    }
    \subfloat[\textit{GNN-Max}]{
        \includegraphics[width=0.3\textwidth]{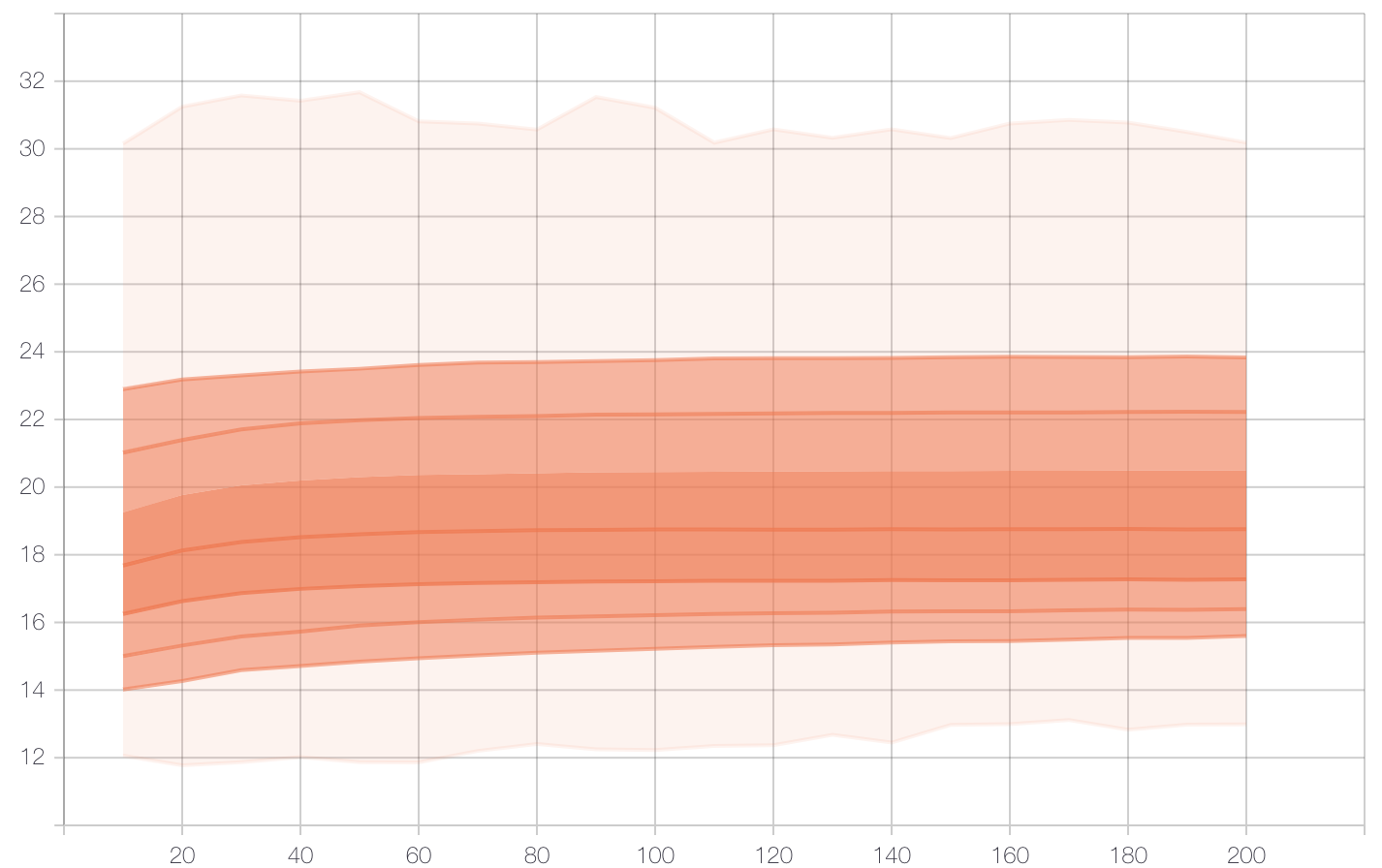}
        \label{fig:gnn-max_emb_2norm}
    }
    \subfloat[\textit{GNN-Max + learnt BN}]{
        \includegraphics[width=0.3\textwidth]{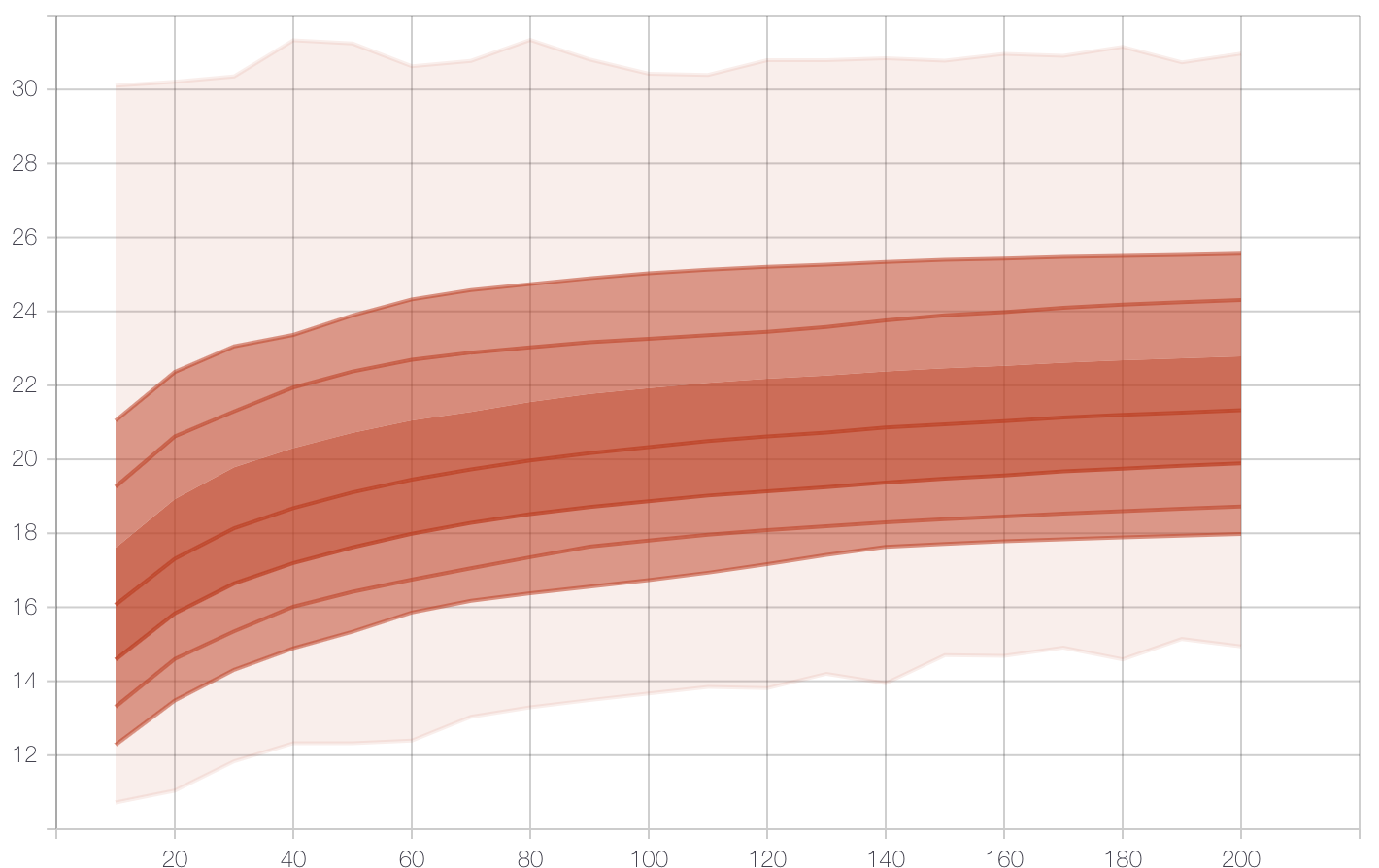}
        \label{fig:gnn-max-bntrack_emb_2norm}
    }
    \caption{\small Distribution plots of \textbf{node} embedding $\ell_2$ norms (y-axis) across TSP sizes (x-axis).
	}
	\label{fig:emb_2norm}
\end{figure}


\begin{figure}[t!]
	\centering
	\subfloat[\textit{GNN-Sum}]{
        \includegraphics[width=0.3\textwidth]{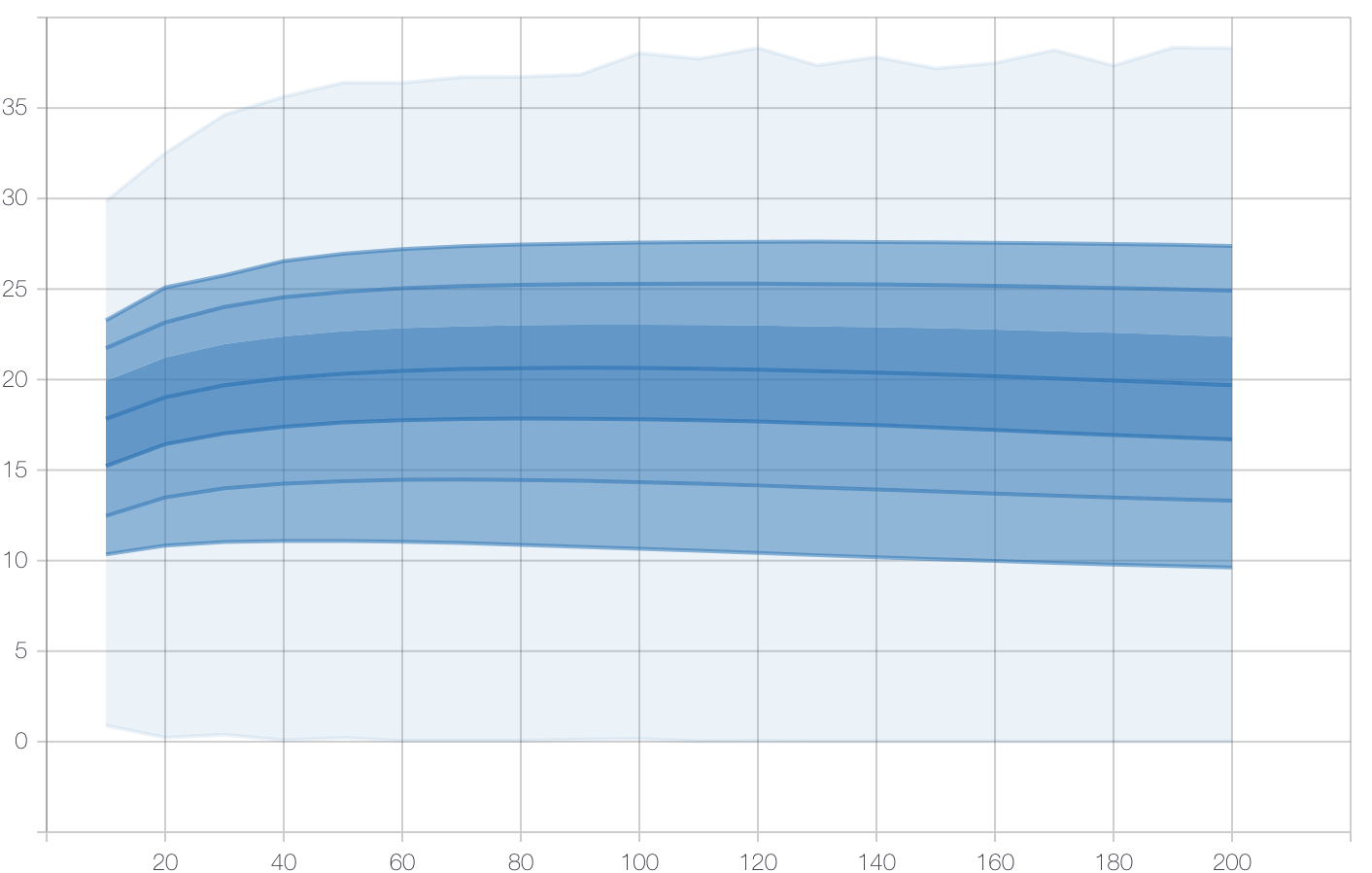}
        \label{fig:gnn-sum_emb_dist}
    }
    \subfloat[\textit{GNN-Max}]{
        \includegraphics[width=0.3\textwidth]{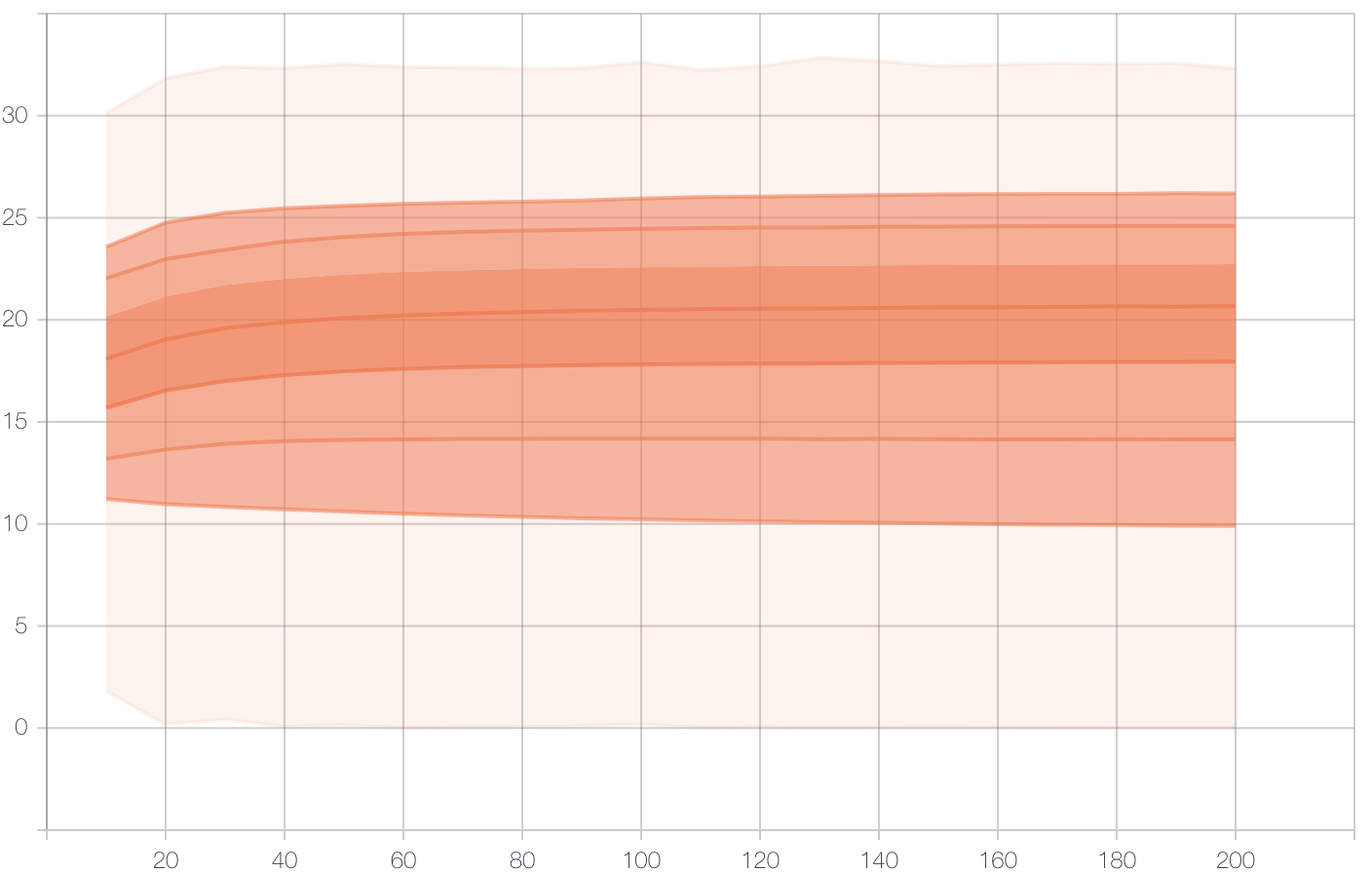}
        \label{fig:gnn-max_emb_dist}
    }
    \subfloat[\textit{GNN-Max + learnt BN}]{
        \includegraphics[width=0.3\textwidth]{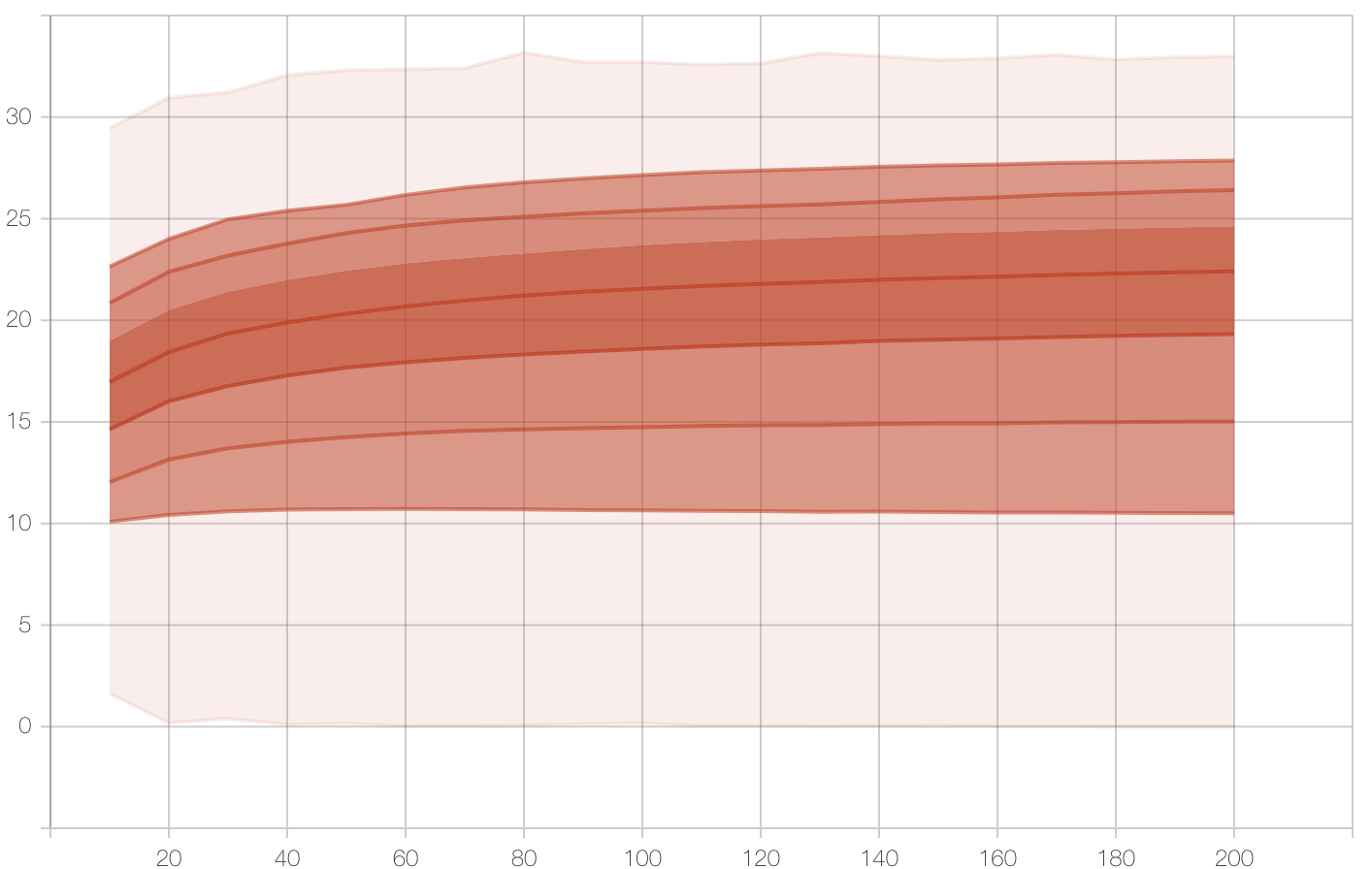}
        \label{fig:gnn-max-bntrack_emb_dist}
    }
    \caption{\small Distribution plots of \textbf{node} embedding pair-wise distances (y-axis) across TSP sizes (x-axis).
	}
	\label{fig:emb_dist}
\end{figure}



\textbf{Node Embedding Space\quad}
In Figures~\ref{fig:emb_2norm} and \ref{fig:emb_dist}, we see that \textit{GNN-Max} leads to the most stable node embedding norms and pair-wise distances (which are calculated at an intra-graph level) across TSP sizes.
On the other hand, \textit{GNN-Sum} and \textit{GNN-Max + learnt BN} lead to fluctuating and monotonically increasing embedding norms as size increases, \textit{e.g.} compare Figure~\ref{fig:gnn-max_emb_2norm} and Figure~\ref{fig:gnn-max-bntrack_emb_2norm}.
Clearly, maintaining similar distributions for node embeddings across graph sizes indicates that the GNN is building meaningful representations of local structure, or, at the very least, does not break down for large graphs.
This enables better generalization, as the decoder has lower chances of encountering embeddings which are statistically different than those seen during training.



\begin{figure}[t!]
	\centering
	\subfloat[\textit{GNN-Sum}]{
        \includegraphics[width=0.3\textwidth]{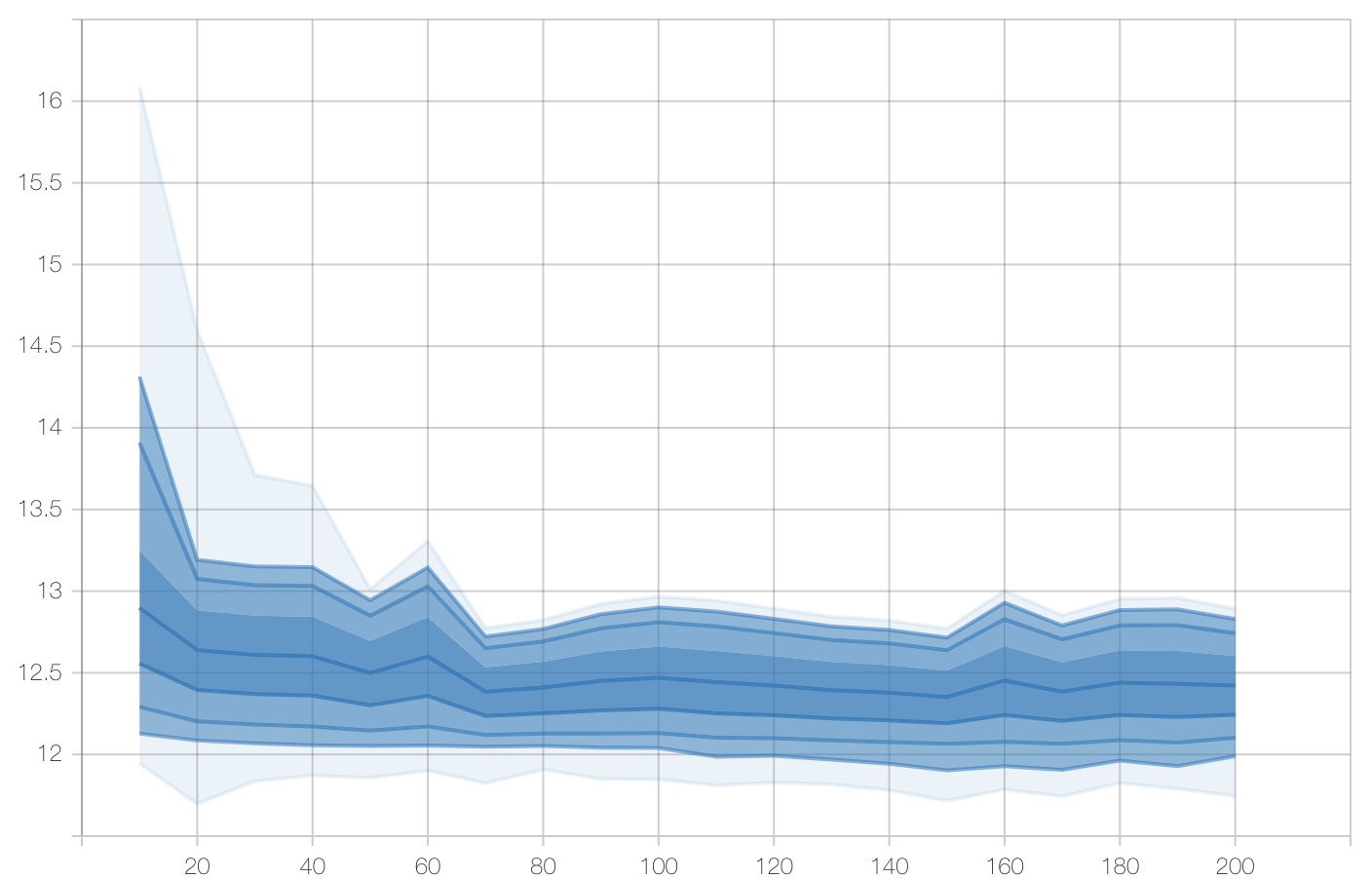}
        \label{fig:gnn-sum_gemb_2norm}
    }
    \subfloat[\textit{GNN-Max}]{
        \includegraphics[width=0.3\textwidth]{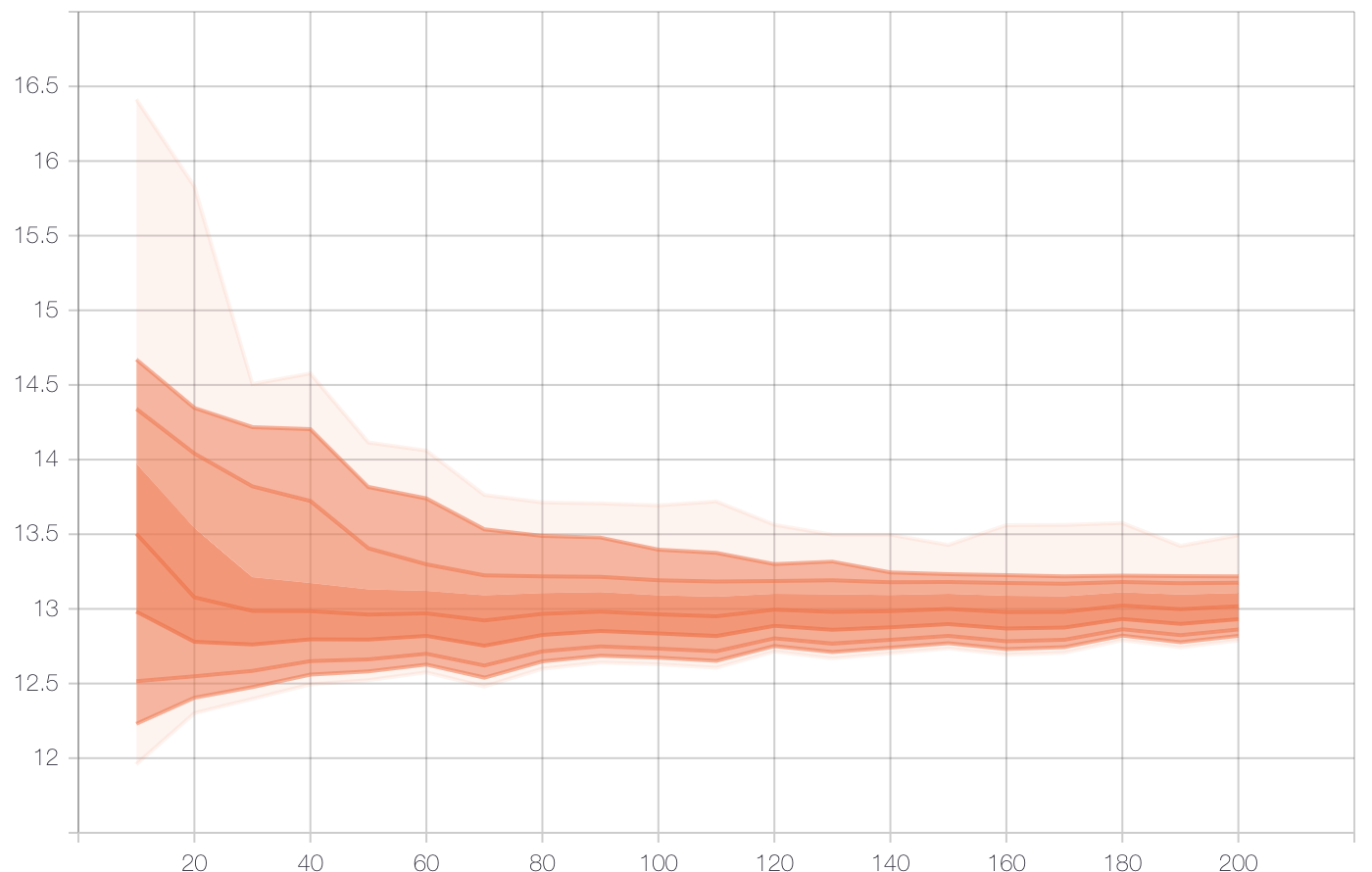}
        \label{fig:gnn-max_gemb_2norm}
    }
    \subfloat[\textit{GNN-Max + learnt BN}]{
        \includegraphics[width=0.3\textwidth]{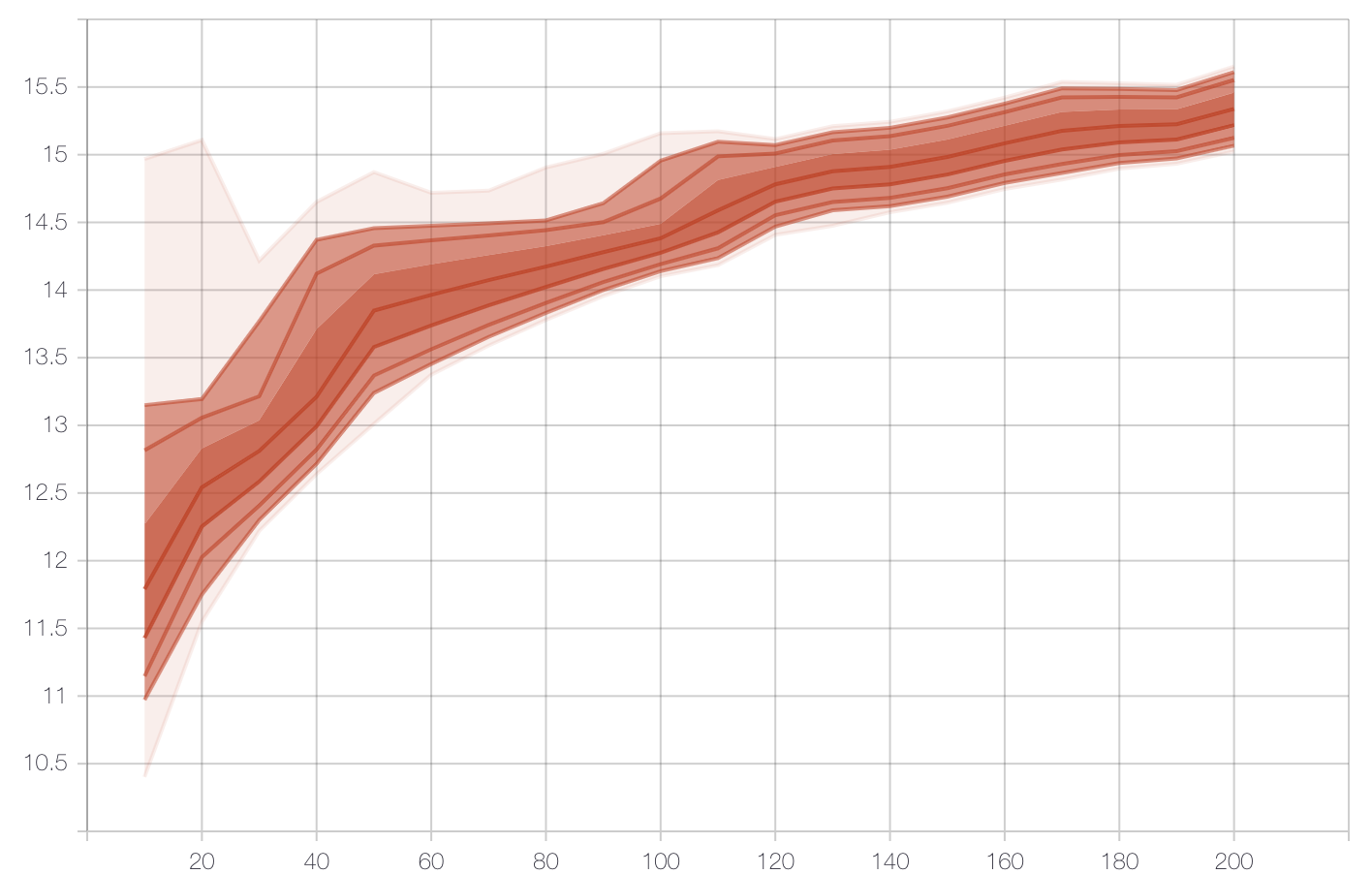}
        \label{fig:gnn-max-bntrack_gemb_2norm}
    }
    \caption{\small Distribution plots of \textbf{graph} embedding $\ell_2$ norms (y-axis) across TSP sizes (x-axis).
	}
	\label{fig:gemb_2norm}
\end{figure}


\begin{figure}[t!]
	\centering
	\subfloat[\textit{GNN-Sum}]{
        \includegraphics[width=0.3\textwidth]{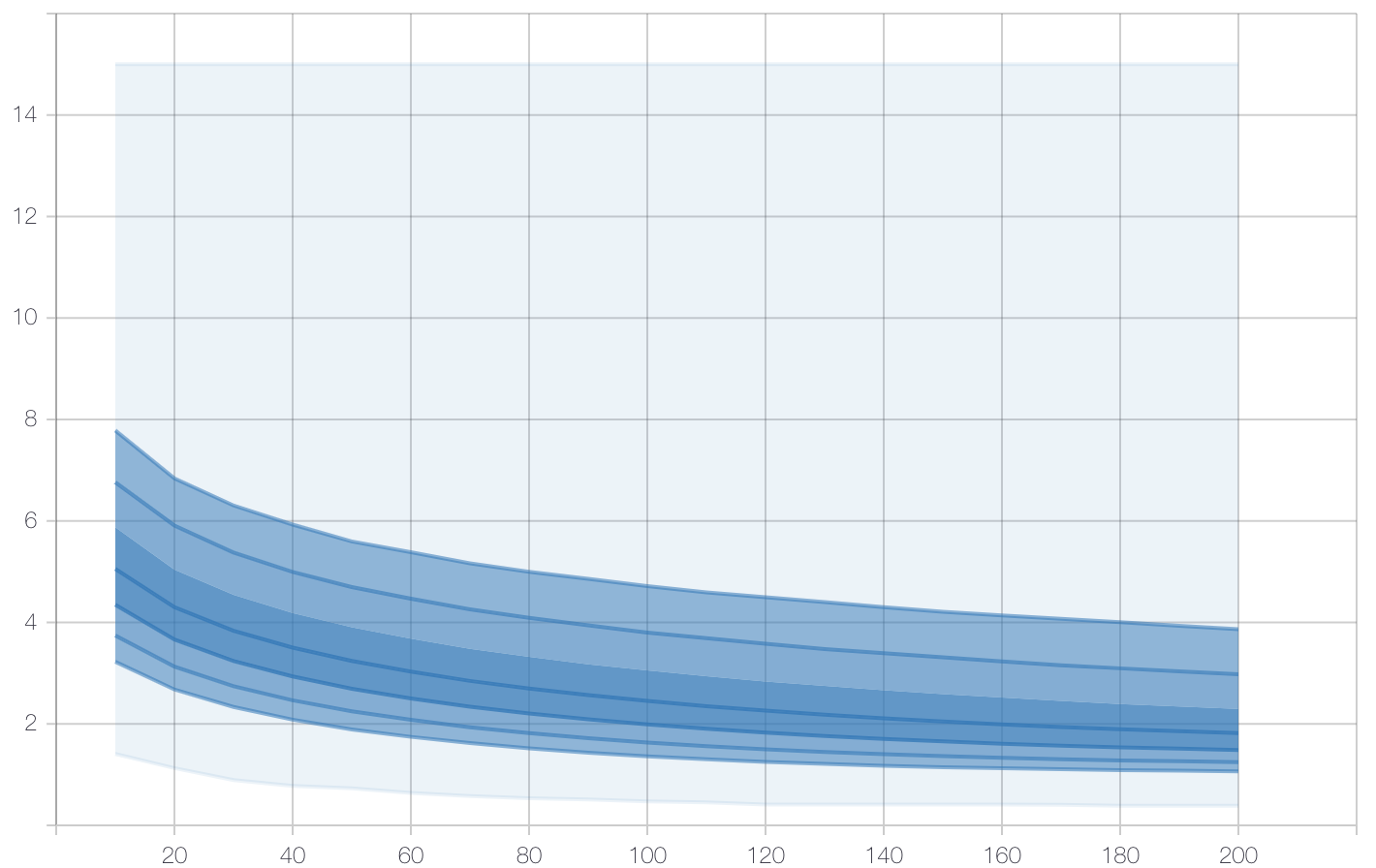}
        \label{fig:gnn-sum_gemb_dist}
    }
    \subfloat[\textit{GNN-Max}]{
        \includegraphics[width=0.3\textwidth]{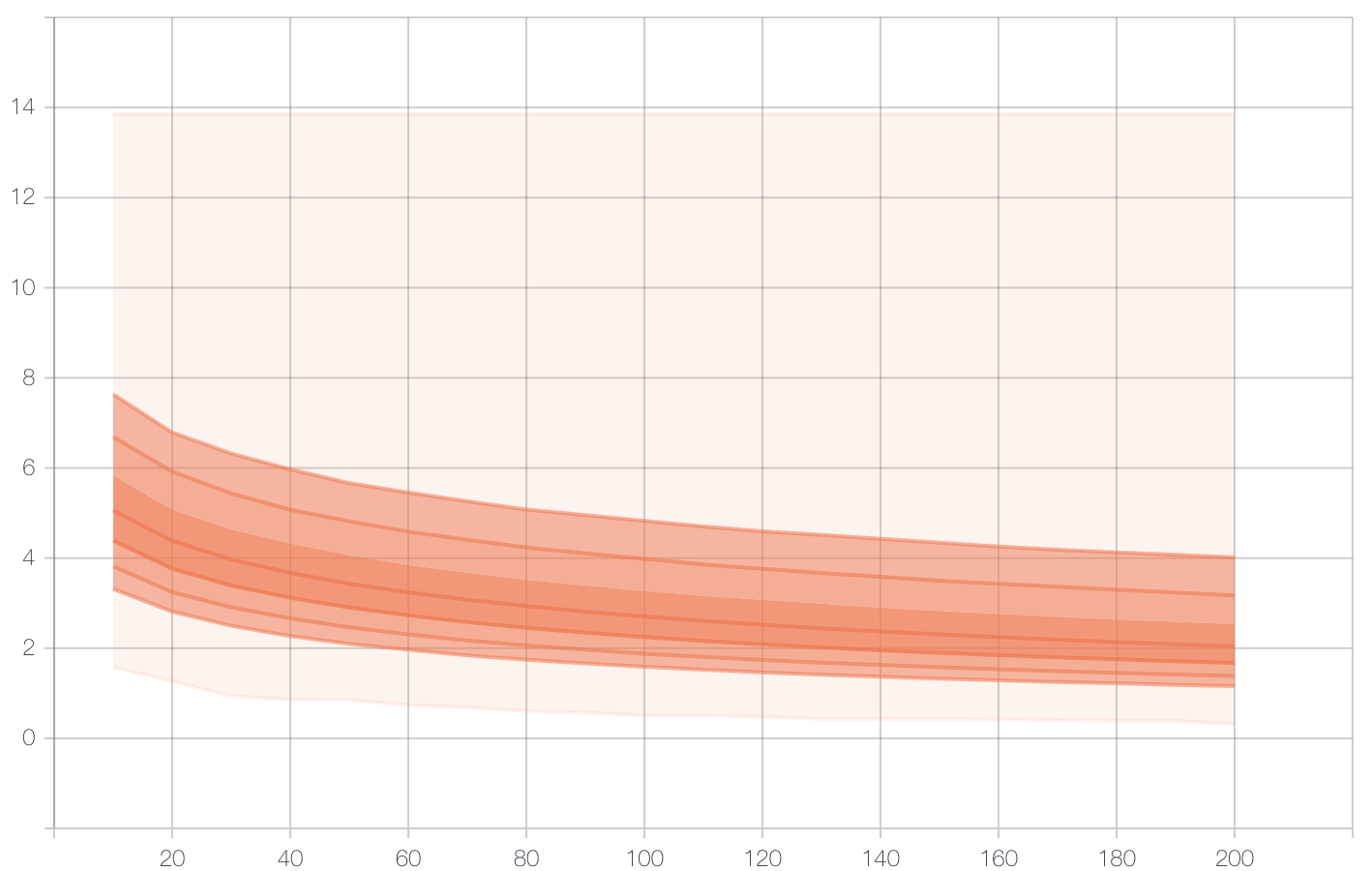}
        \label{fig:gnn-max_gemb_dist}
    }
    \subfloat[\textit{GNN-Max + learnt BN}]{
        \includegraphics[width=0.3\textwidth]{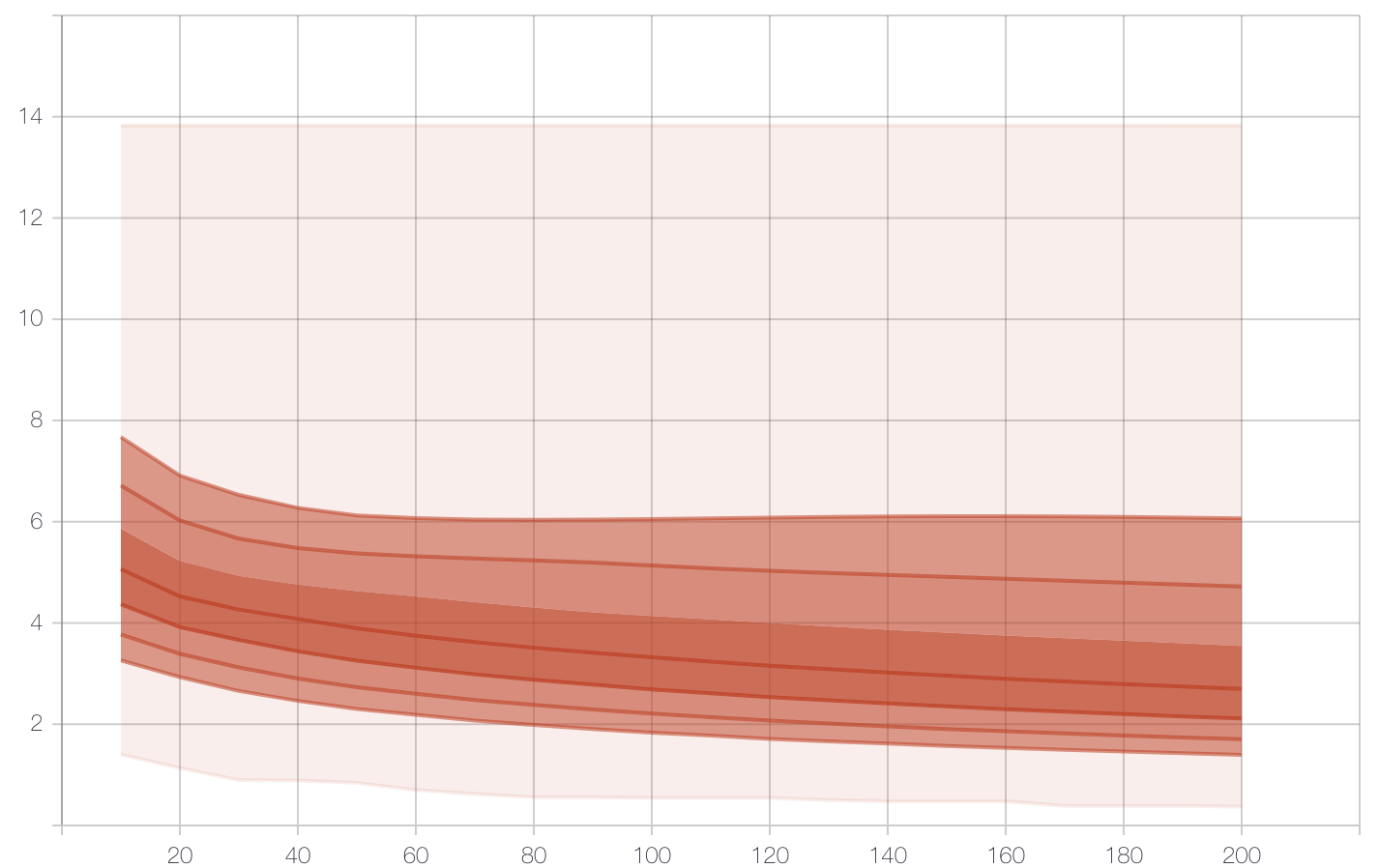}
        \label{fig:gnn-max-bntrack_gemb_dist}
    }
    \caption{\small Distribution plots of \textbf{graph} embedding pair-wise distances (y-axis) across TSP sizes (x-axis).
	}
	\label{fig:gemb_dist}
\end{figure}


\textbf{Graph Embedding Space\quad}
Figures~\ref{fig:gemb_2norm} and \ref{fig:gemb_dist} indicate that the graph embedding space is shrinking towards a single magnitude and moving closer as graph size increases.
Interestingly, with standard BatchNorm, the graph embedding magnitude monotonically increases with graph size to ranges beyond those for training graphs.
On the other hand, using batch statistics for BatchNorm, as done in \textit{GNN-Max} and \textit{GNN-Sum}, leads to graph embedding magnitudes converging to a single value which is within the range of values for training graphs, thus enabling better generalization.
\textit{E.g.} compare Figure~\ref{fig:gnn-max_gemb_2norm} and Figure~\ref{fig:gnn-max-bntrack_gemb_2norm}.

\begin{figure}[t!]
	\centering
	\subfloat[\textit{GNN-Max}]{
        \includegraphics[width=0.45\textwidth]{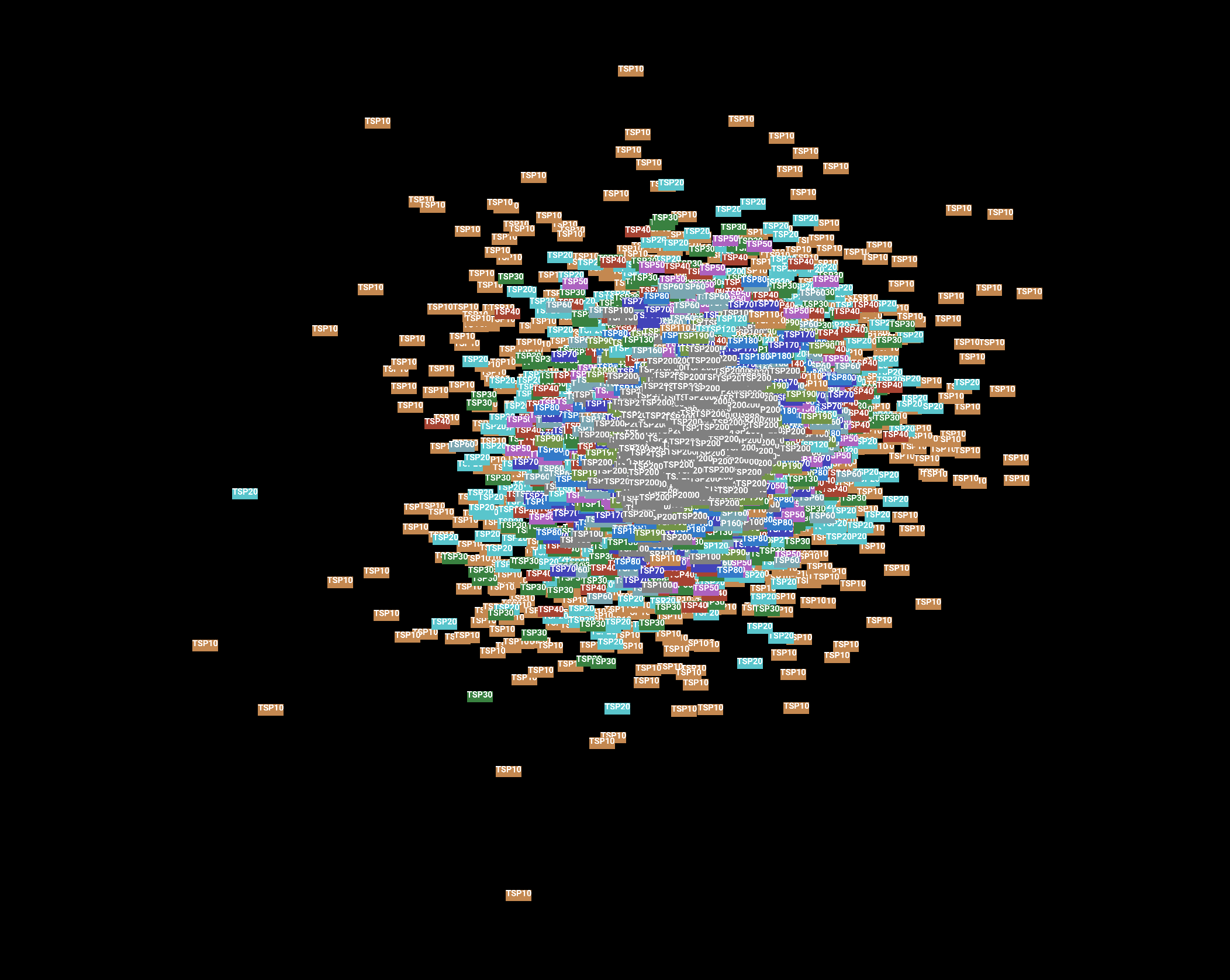}
        \label{fig:gnn-max_pca}
    }
    \quad
    \subfloat[\textit{GNN-Max + learnt BN}]{
        \includegraphics[width=0.45\textwidth]{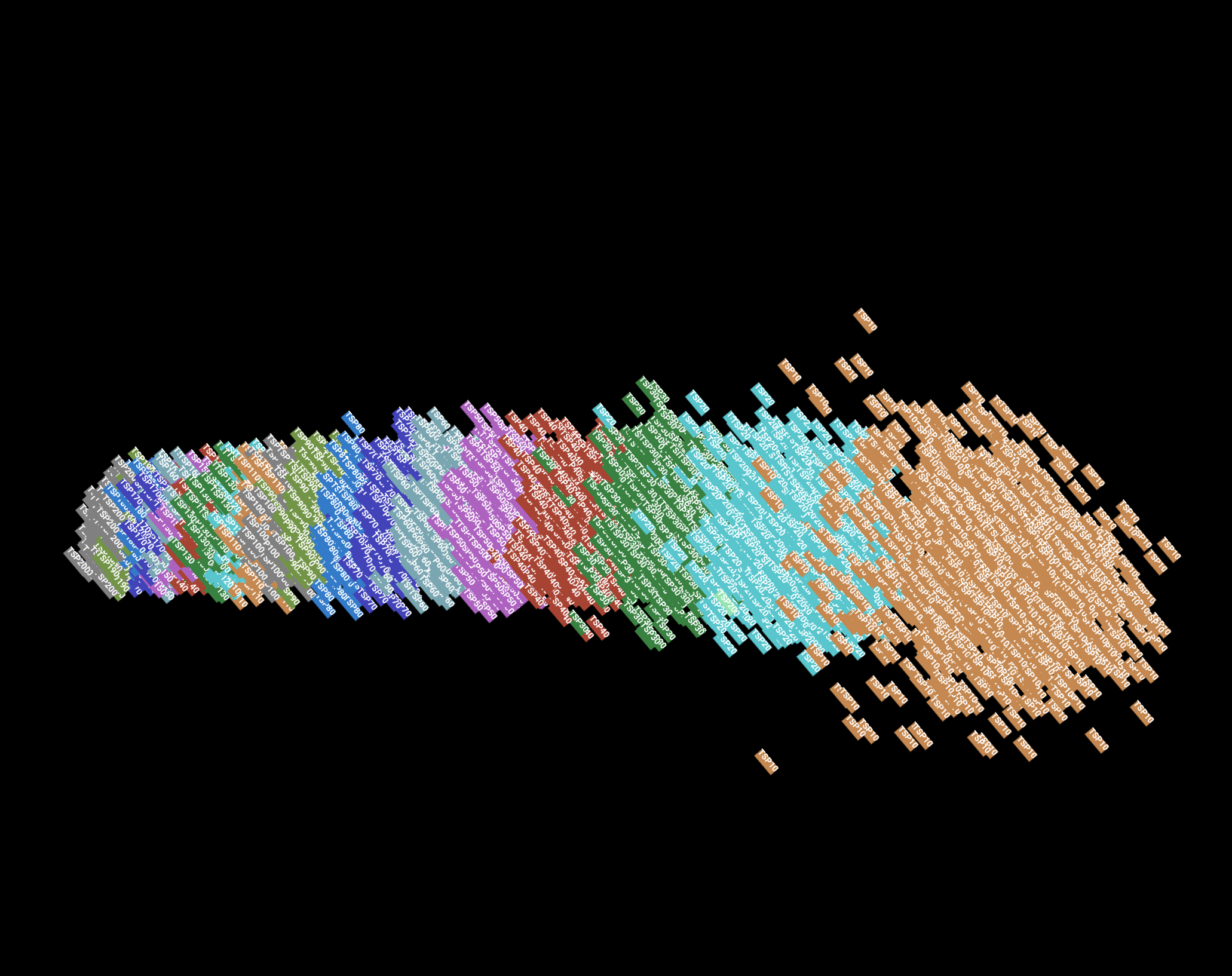}
        \label{fig:gnn-max-bntrack_pca}
    }
	\label{fig:pca}
	\caption{\small 2D PCA of graph embedding spaces. Colors represent TSP instance sizes, \textit{e.g.} orange: TSP10, teal: TSP20, pink: TSP50, dark grey: TSP200.
	}
\end{figure}

We can further visualize this phenomenon through 2D Principal Component Analysis (PCA) plots of graph embedding spaces for \textit{GNN-Max} and \textit{GNN-Max + learnt BN} models, see Figures~\ref{fig:gnn-max_pca} and \ref{fig:gnn-max-bntrack_pca}.
In both cases, the graph embeddings at larger sizes have very similar magnitudes and are extremely close to each other, indicating that the model is unable to differentiate among different graphs.
Thus, decoders currently lack good global structural context.
Investigating better graph embeddings through pooling methods~\cite{ying2018hierarchical} could be an interesting approach towards representing global graph structure beyond training sizes.

\section{Extra Results}
\label{app:omitted}


\begin{figure}[t!]
\begin{minipage}{.48\textwidth}
\centering
    \includegraphics[width=0.98\linewidth]{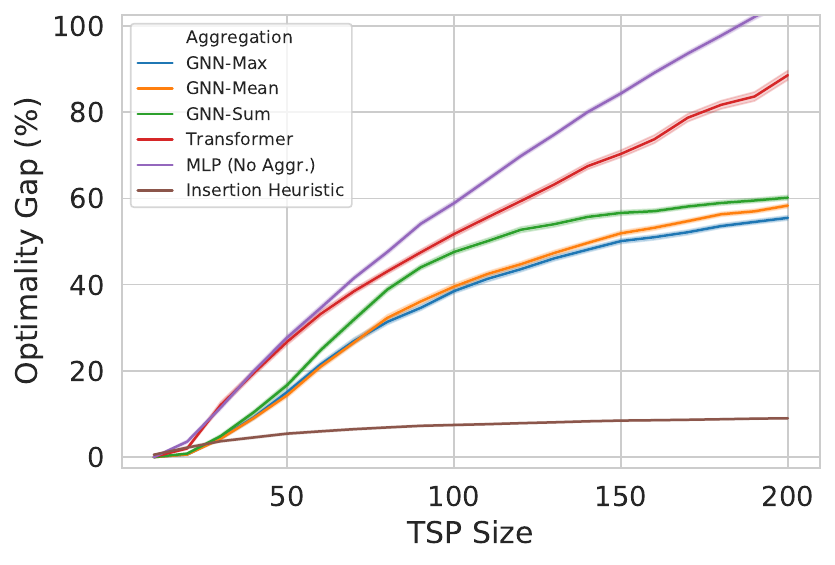}
    \caption{\small GNN aggregation functions (NAR decoder).}
    \label{fig:gnn_aggregation-nar}
\end{minipage}
\quad
\begin{minipage}{.48\textwidth}
\centering
    \includegraphics[width=0.98\linewidth]{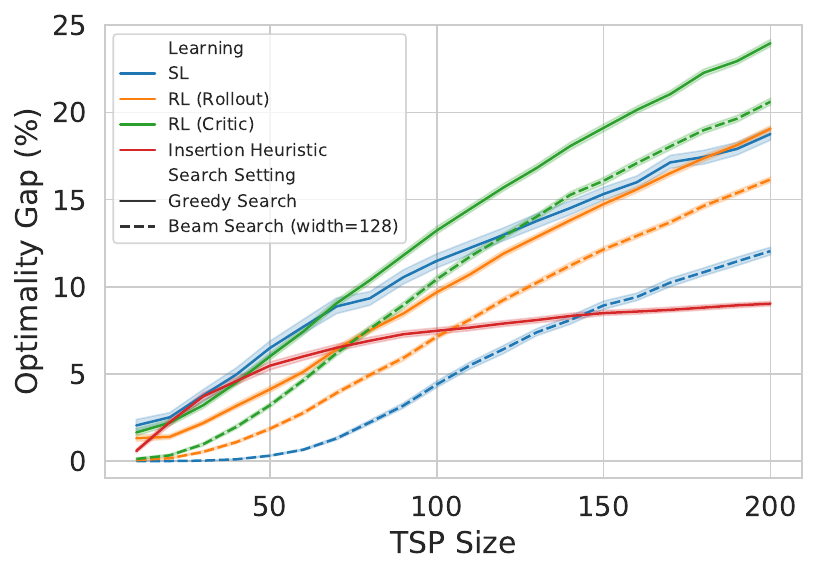}
    \caption{\small Comparing learning paradigms and solution search settings.}
    \label{fig:learning_and_search_critic}
\end{minipage}
\end{figure}



\textbf{NAR Decoders and Aggregation Functions\quad}
In Section~\ref{sec:experiments}, we found that AR decoding provides a powerful sequential inductive bias for TSP and is able to generalizes well with both GNNs as well as structure-agnostic encoder architectures.
This result may lead one to question the need for GNNs, altogether.
Interestingly, Figure~\ref{fig:gnn_aggregation-nar} illustrates a different trend for NAR architectures:
GNN encoders generalize better than both Transformers and MLPs, indicating that leveraging graph structure is essential in the absence of the sequential inductive bias.
(It is worth noting that, overall, all models with NAR decoders generalize poorly compared to AR architectures for our experimental setup.)

\textbf{Critic baseline\quad}
Figure~\ref{fig:learning_and_search_critic} illustrates that, for identical models, the critic baseline~\cite{bello2016neural,deudon2018learning} is unable to match the performance of the rollout baseline~\cite{kool2018attention} under both greedy and beam search settings.
We did not explore tuning learning rates and hyperparameters for the critic network, opting to use the same settings as those for the actor.
In general, getting actor-critic methods to work seems to require more parameter tuning than the rollout baseline.





\begin{figure}[t!]
\centering
\begin{minipage}{.46\textwidth}
\centering
    \includegraphics[width=\linewidth]{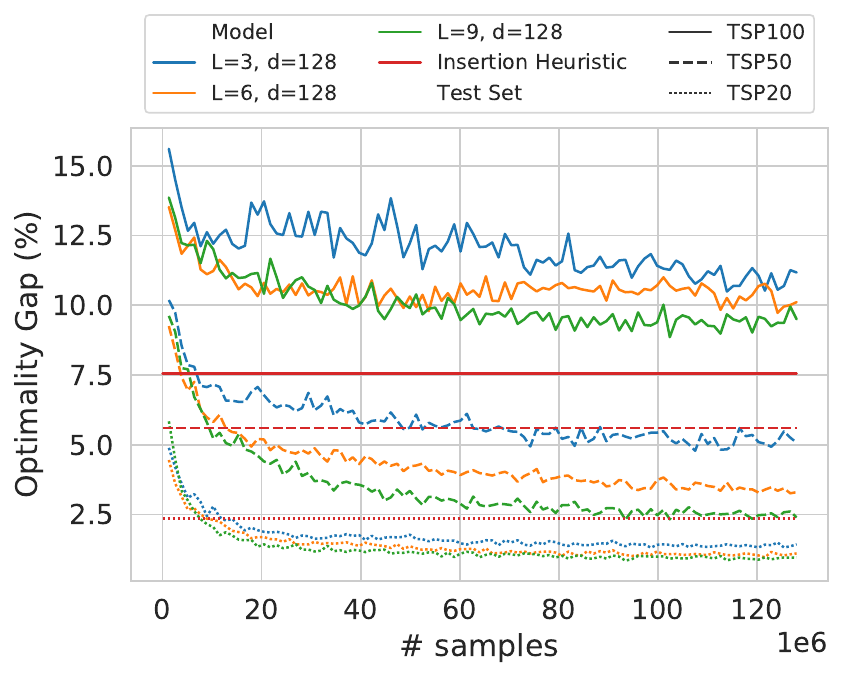}
    \caption{\small Scaling computation and model parameters for AR decoder.}
    \label{fig:scale-ar-sl}
\end{minipage}
\quad
\begin{minipage}{.46\textwidth}
\centering
    \includegraphics[width=\linewidth]{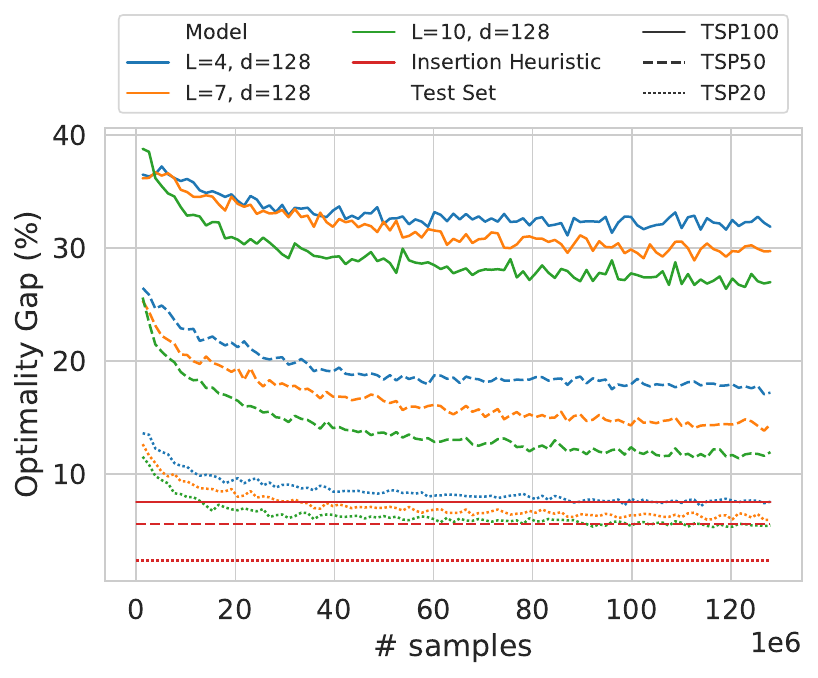}
    \caption{\small Scaling computation and model parameters for NAR decoder.}
    \label{fig:scale-ar-sl-nar}
\end{minipage}
\end{figure}

\textbf{Scaling computation for AR and NAR architectures\quad}
In Figures~\ref{fig:scale-ar-sl} and \ref{fig:scale-ar-sl-nar}, we present extended results for Section~\ref{sec:experiments:scale}, where we scale model parameters and data.
We observe that using larger models (up to 1.5~Million parameters) enables fitting the training dataset better.
The impact of larger models is especially evident for NAR architectures.
As previously noted, recent NAR-based models~\cite{nowak2017note,joshi2019efficient} used more than 30 layers with over 10~Million parameters to outperform AR architectures on fixed TSP sizes.
We believe that such overparameterized networks are able to memorize all patterns for small TSP training sizes~\cite{zhang2016understanding},
but the learnt policy is unable to generalize beyond training graph sizes as NAR decoding does not provide a useful inductive bias for TSP. 


\section{Visualizing Model Predictions}
\label{app:viz}
\begin{figure}[ht!]
    \centering
    \includegraphics[width=0.8\linewidth]{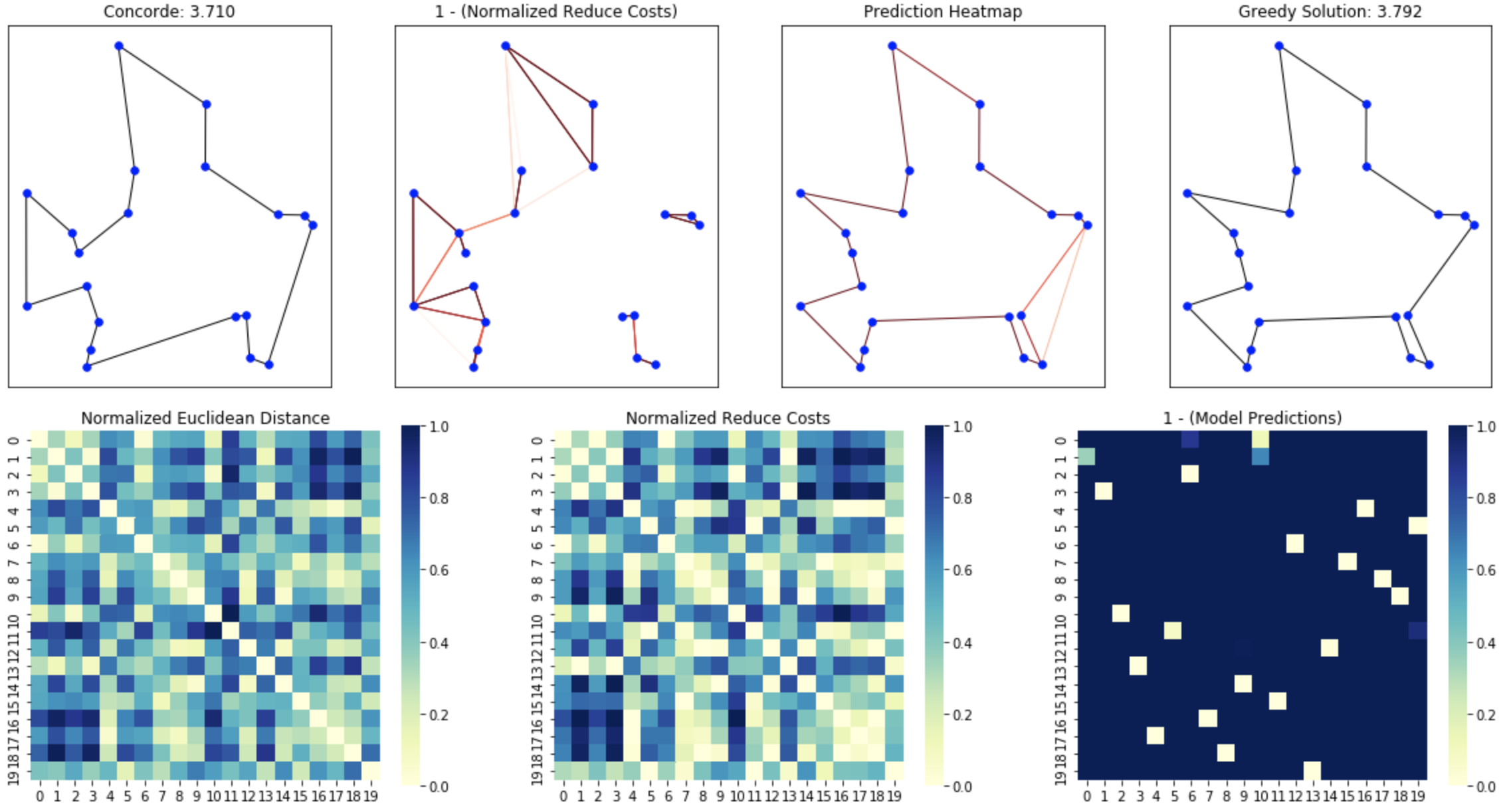}
    \caption{\small Prediction visualization for TSP20.}
    \label{fig:viz_tsp20}
\end{figure}

\begin{figure}[ht!]
    \centering
    \includegraphics[width=0.8\linewidth]{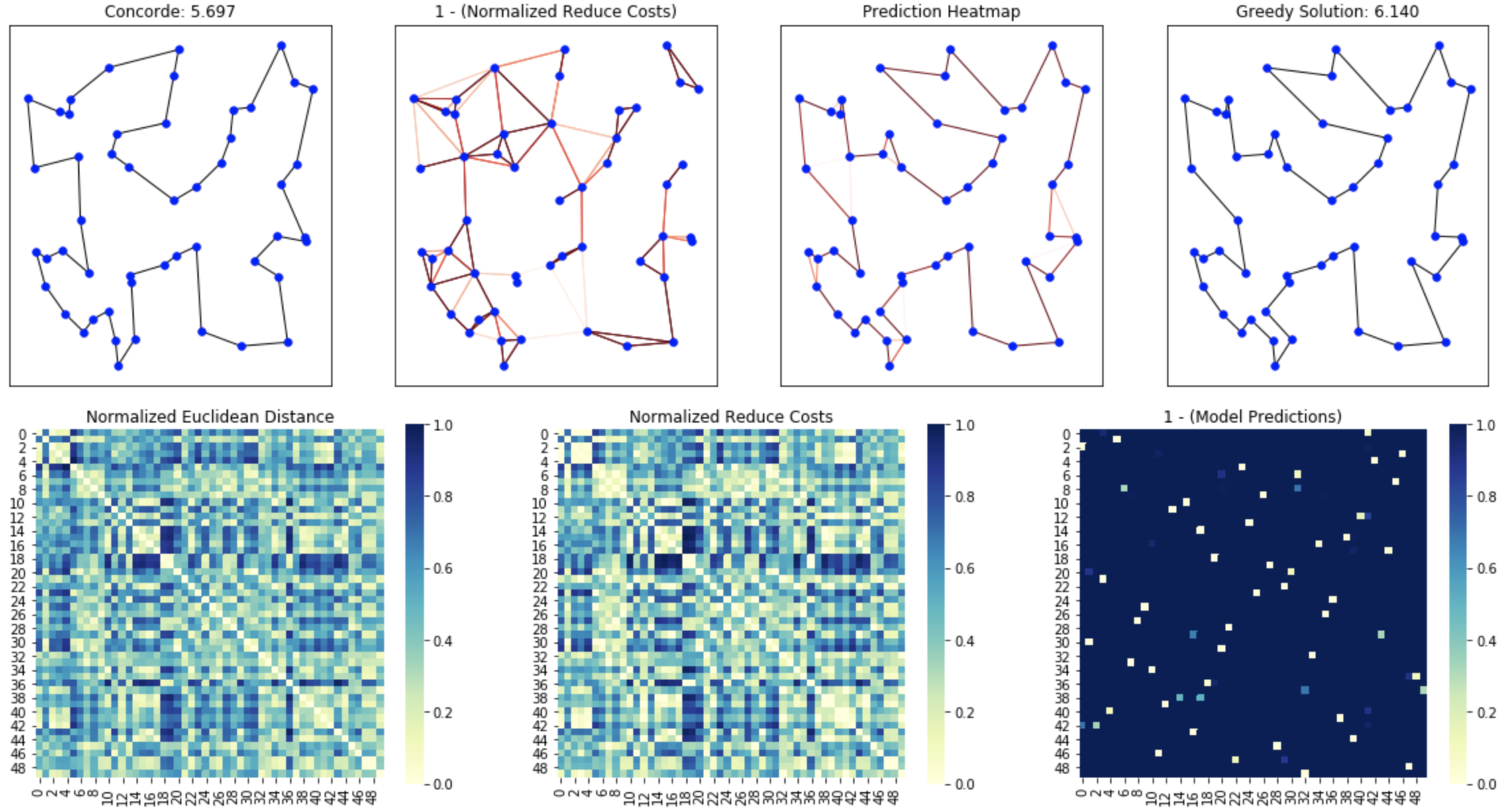}
    \caption{\small Prediction visualization for TSP50}
    \label{fig:viz_tsp50}
\end{figure}

As a final note, we present a visualization tool for generating model predictions and heatmaps of TSP instances, see Figures~\ref{fig:viz_tsp20}, \ref{fig:viz_tsp50}. 
We advocate for the development of more principled approaches to neural combinatorial optimization, \textit{e.g.} along with model predictions, visualizing the reduce costs for each edge (cheaply obtained using the Gurobi solver~\cite{gurobi2015gurobi}) may help debug and improve learning-driven approaches in the future.
Using reduce costs as supervision signals could also be an inexpensive alternative to running optimal solvers to create large labelled datasets.



\end{document}